\documentclass[10pt,twocolumn,letterpaper]{article}

\usepackage[pagenumbers]{cvpr} 

\definecolor{cvprblue1}{rgb}{0.21,0.49,0.74}
\usepackage[pagebackref,breaklinks,colorlinks,allcolors=cvprblue1]{hyperref}

\usepackage{kotex}
\usepackage{xcolor}

\usepackage{pifont}
\usepackage{fontawesome5}
\usepackage{svg}
\usepackage[utf8]{inputenc}
\usepackage{array,multirow,graphicx}
\usepackage{float}
\usepackage{amsmath}
\usepackage{makecell}
\usepackage{mathtools}
\usepackage{bbm}
\usepackage{hhline}
\usepackage{boldline}
\usepackage{booktabs}
\usepackage{multirow}
\usepackage{multicol}
\usepackage{algorithm} 
\usepackage{algpseudocode} 
\usepackage{graphicx}
\usepackage{amssymb}
\usepackage[dvipsnames]{xcolor}
\usepackage{colortbl}
\usepackage{arydshln}
\usepackage{booktabs} 
\usepackage[accsupp]{axessibility} 
\usepackage[table]{xcolor} 
\usepackage[most]{tcolorbox}
\usepackage{wrapfig}
\newcommand{\myxmark}{\ding{55}}
\newcommand{\mychecking}{\ding{51}}
\usepackage[ruled,vlined,algo2e]{algorithm2e}
\usepackage{listings}
\usepackage{amsfonts}
\usepackage{placeins}
\definecolor{headercolor}{gray}{0.92}
\definecolor{customblue}{RGB}{75,75,225}
\definecolor{maroon}{cmyk}{0,0.87,0.68,0.32}

\definecolor{cvprblue}{HTML}{d0ebff}
\definecolor{mypink}{HTML}{ffc9c9}
\definecolor{myred}{HTML}{f03e3e}

\usepackage{amssymb}
\usepackage{pifont}
\usepackage{etoc}

\newcommand{\ourmodel}{\textsc{ViKey}}

\def\paperID{3690} 
\def\confName{CVPR}
\def\confYear{2026}

\title{\ourmodel: Enhancing Temporal Understanding in Videos via Visual Prompting}

\author{
Yeonkyung Lee$^{*}$ \hspace{0.6em}
Dayun Ju$^{*}$ \hspace{0.6em}
Youngmin Kim \hspace{0.6em}
Seil Kang \hspace{0.6em}
Seong Jae Hwang\\
Yonsei University\\
{\tt\small \{yeonkyung.lee, juda0707, winston1214, seil, seongjae\}@yonsei.ac.kr}\\
}

\begin{document}
\maketitle

\begin{abstract}
Recent advancements in Video Large Language Models (VideoLLMs) have enabled strong performance across diverse multimodal video tasks. To reduce the high computational cost of processing dense video frames, efficiency-oriented methods such as frame selection have been widely adopted. While effective at minimizing redundancy, these methods often cause notable performance drops on tasks requiring temporal reasoning. Unlike humans, who can infer event progression from sparse visual cues, VideoLLMs frequently misinterpret temporal relations when intermediate frames are omitted. To address this limitation, we explore visual prompting (VP) as a lightweight yet effective way to enhance temporal understanding in VideoLLMs. Our analysis reveals that simply annotating each frame with explicit ordinal information helps the model perceive temporal continuity. This visual cue also supports frame-level referencing and mitigates positional ambiguity within a sparsely sampled sequence. Building on these insights, we introduce \ourmodel, a training-free framework that combines VP with a lightweight Keyword–Frame Mapping (KFM) module. KFM leverages frame indices as dictionary-like keys to link textual cues to the most relevant frames, providing explicit temporal anchors during inference. Despite its simplicity, our approach substantially improves temporal reasoning and, on some datasets, preserves dense-frame baseline performance with as few as 20\% of frames.
\begingroup
\renewcommand{\thefootnote}{}
\footnote{\faGithub\hspace{0.3em}Code: \href{https://github.com/MICV-yonsei/ViKey}{https://github.com/MICV-yonsei/ViKey}}
\footnote{* Equal contribution}
\endgroup
\end{abstract}
    
%
\section{Introduction}
\begin{figure}[ht!]
  \centering
  \includegraphics[width=\linewidth]{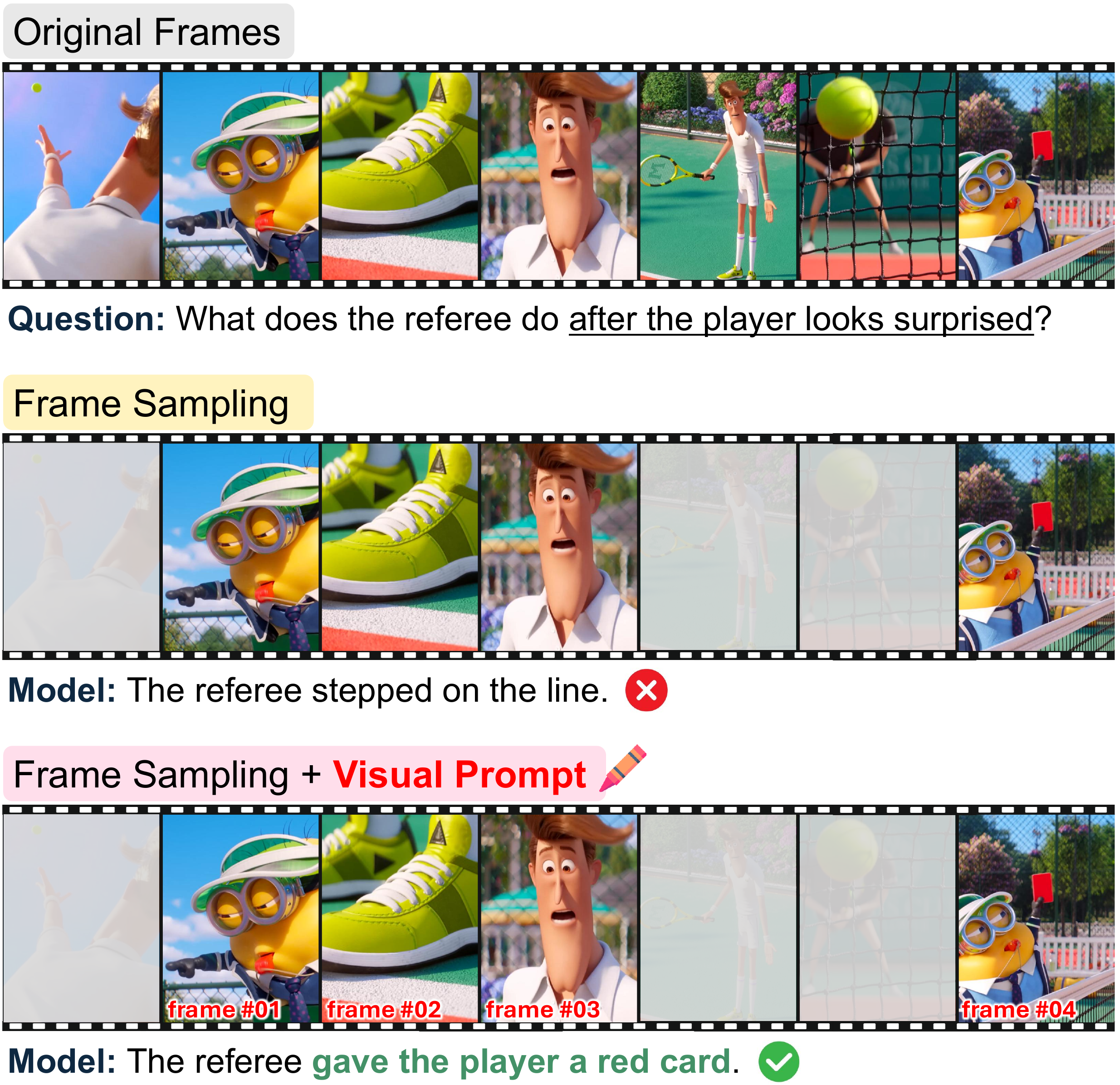}
  \vspace{-18 pt}
  \caption{\textbf{Illustrations of How Frame Selection Disrupts and Visual Prompting Restores Temporal Reasoning.} Frame selection improves efficiency but breaks temporal continuity, causing VideoLLMs to misinterpret transitions. With frame-index visual prompts, the model regains temporal order and correctly identifies causal relations under sparse-frame conditions.}
  \label{fig:fig_1}
\vspace{-10 pt}
\end{figure}
%
The rapid progress in Video Large Language Models (VideoLLMs) has recently brought significant attention to a range of multimodal tasks~\cite{wang2025internvl3, hong2025glm, comanici2025gemini, bai2025qwen2, zohar2025apollo, zhang2024video}. However, videos consist of long sequences of frames, and processing every frame is computationally and memory-prohibitive. Consequently, frame selection, which removes redundant frames and preserves only salient moments, has become a practically indispensable component of VideoLLM pipelines.

Despite improving overall computational efficiency, frame selection strategies often lead to non-negligible performance degradation. \cref{fig:fig_1} illustrates this phenomenon. The example video shows a player stepping over the line and the referee giving that player a red card. Even when only a few sampled frames are provided, a human observer can readily infer the sequence of events. In contrast, once intermediate frames are removed, a VideoLLM fails to capture this key temporal relation and instead incorrectly concludes that the referee stepped on the line. These failures are particularly pronounced in temporal understanding tasks, even when sufficient visual evidence remains in the input.

One key factor in this limitation is that frame selection tends to interrupt the continuous flow of visual information, offering the model only a set of temporally discrete frames~\cite{longvideo_temporalreasoning}. From the model’s perspective, it no longer receives smoothly connected observations along the time axis, as in the original video. Under such settings, reconstructing a temporally coherent sequence of events from the fragmentary cues contained in individual frames is fundamentally difficult. 
To address this challenge, recent studies have investigated methods such as enhanced temporal encoding, extended context modules, and explicit spatio-temporal modelling~\cite{transfer_t2v,longvideo_temporalreasoning,vt_temporalsequence,step}. While these methods have achieved meaningful progress, they often involve complex design changes, extensive training data or supervision.

In search of a more efficient alternative, we turned our attention to visual prompting (VP). VP has emerged as a lightweight yet effective mechanism for steering model behaviour without modifying network architectures or requiring additional supervision. Prior studies~\cite{VP, videoVP} show that simple visual cues, such as highlighting regions, drawing circles, or overlaying spatial markers, can substantially improve visual grounding and guide spatial attention within individual frames. However, these techniques have been explored almost exclusively in spatial settings within single images or frames. Their potential to serve as linking cues across frames, and thereby support temporal coherence and reasoning in videos, remains largely underexplored.

Building on these observations, we propose \ourmodel, a simple approach that enhances the temporal reasoning capability of VideoLLMs. As illustrated in \cref{fig:fig_1}, we introduce explicit frame-index prompts (\eg, ``frame \#01'') over each input frame, allowing the model to perceive temporal order across sparsely sampled sequences.

To systematically analyze how such visual prompts affect temporal reasoning, we design three experiments:

\begin{enumerate}
    \item \textbf{Positional Embedding Degradation:} We disrupt the model’s temporal positional encoding to test whether visual prompts alone can restore the sense of frame order.
    
    \item \textbf{Frame-level Referencing in VideoLLMs:} We investigate whether VideoLLMs can treat frame numbers as dictionary-like keys for referencing visual content. We then analyze how the spatial placement of visual prompts affects this capability and reveal a strong positional bias.
    
    \item \textbf{Attention-based Analysis:} We analyze layer-wise attention patterns to determine whether visual prompts promote stronger temporal alignment between visual and textual representations.
\end{enumerate} 

Motivated by these analyses, we introduce numbered visual prompts as a lightweight Keyword-Frame Mapping (KFM) module. KFM treats frame indices as dictionary-like keys and links textual cues to frames via simple similarity in a shared embedding space.
For each query, salient keywords are matched to the most relevant frame and the query is reformulated with the mapped index, providing explicit temporal anchors.

Through extensive experiments on temporal-reasoning benchmarks, we show that our approach reliably strengthens VideoLLM performance. Even with only 20\% of frames, adding visual prompts preserves accuracy comparable to a dense-frame baseline on some datasets, demonstrating robustness under sparse inputs. Beyond these gains, we analyze optimized VP characteristics and their interaction with KFM. By combining VP with this lightweight mapping, \ourmodel{} delivers a training-free, plug-and-play enhancement that balances accuracy and efficiency and transfers across diverse VideoLLMs and video tasks.
\section{Related Works}
\label{sec:related_works}
%
\begin{figure*}[t!]
  \centering
  \includegraphics[width=\linewidth]{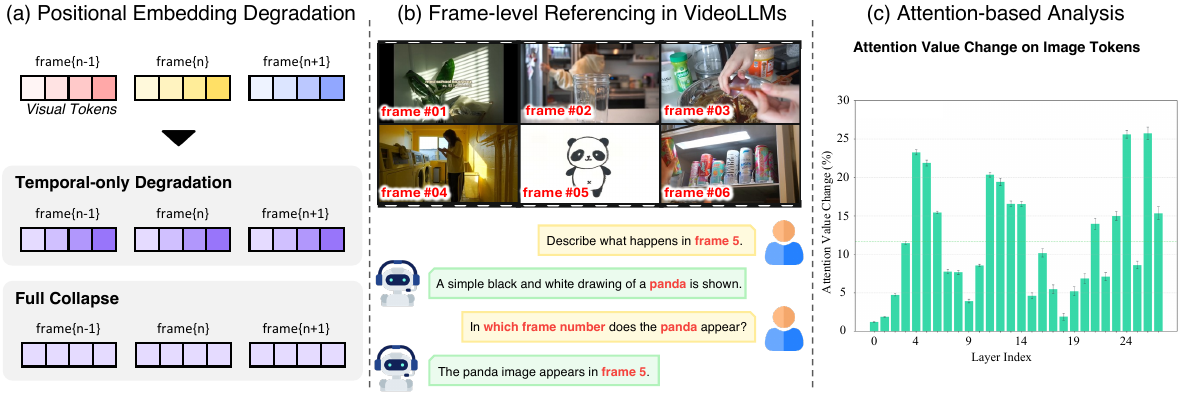}
  \vspace{-18 pt}
  \caption{
   \textbf{Probing VP for temporal understanding.} (a) Positional Embedding Degradation: remove temporal order only, or collapse both temporal and spatial positions. (b) Frame-level Referencing in VideoLLMs: frame-index prompts enable lookup and reverse-lookup between indices and content. (c) Attention Analysis: VP increases attention to image tokens across layers.
  }
  \label{fig:fig_2}
\vspace{-10 pt}
\end{figure*}
%

\noindent \textbf{Efficient Video Understanding in VideoLLMs.}
Recent studies have explored multiple strategies to improve the efficiency of VideoLLMs on video understanding tasks~\cite{llavavideo, internvideo2, qwen2vl, videollama}. 
Token-efficient methods reduce computational cost by compressing or removing redundant visual tokens~\cite{dycoke, llavascissor, framefusion, llavaprumerge}. 
In parallel, frame selection methods choose a subset of representative frames that preserve essential temporal context without processing the entire video sequence~\cite{aks,flexibleframe, samplingdilemma, eff_frame_selection}.

\noindent \textbf{Temporal Reasoning.}
VideoLLMs often lack explicit temporal reasoning mechanisms, limiting their ability to model causal and sequential relations between frames~\cite{jung2025:consistency, maptheflow}. 
Recent benchmarks have systematically evaluated this limitation and shown that current models still struggle with complex temporal reasoning~\cite{vstar, videomme, mlvu, mmbench}. 
To address these challenges, prior work has explored learning causal and order-sensitive relationships between visual and textual modalities~\cite{longvideo_temporalreasoning, causality, timegating}. 
Other approaches enhance grounding and reasoning by leveraging frame order and event boundaries~\cite{vt_temporalsequence, step, order_matter}.
However, most of these methods rely on extensive training or large-scale supervision, leaving temporal reasoning in a training-free and architecture-agnostic setting unsolved.

\noindent \textbf{Visual Prompt Engineering.}
Visual Prompt Engineering injects visual cues directly into the input image while keeping model parameters frozen~\cite{VP}. 
Subsequent work has shown that such prompts can substantially improve accuracy and efficiency in vision–language tasks by helping models capture spatial structure more reliably~\cite{ellipse, emotionVP}. 
Other studies introduce explicit object-level markers, such as strokes or bounding boxes, to strengthen multimodal grounding and reasoning~\cite{fineVP, bboxVP, sterVP}. 
More recently, research has explored user-controllable and interpretable visual prompts~\cite{vipllava, alphaclip}, and has extended these mechanisms from static images to videos to support temporal reasoning and cross-frame consistency~\cite{videoVP, numberit, 3dvp}.
However, most video-based VP methods have primarily treated prompts as training-time signals to improve model performance, without analyzing how or why they work. In contrast, we use VP as a probe to study how VideoLLMs perform temporal reasoning through controlled experiments.

\section{Analyses}
\label{sec:analyses}

To examine the impact of number VP on VideoLLMs, we inserted frame number information (e.g., ``frame \#01'') directly into the pixel space of the image, following the procedure outlined in~\cref{sec:sec_4_2_vp}. Unless otherwise noted, all analyses were conducted using the LLaVA-Video-7B-Qwen~\cite{llavavideo} model. For each of the four datasets used in \cref{sec:5_1}, namely TempCompass~\cite{tempcompass}, MVBench~\cite{mvbench}, VideoMME~\cite{videomme}, and LongVideoBench~\cite{longvideobench}, we randomly sampled 100 examples to construct evaluation subsets, and repeated experiments across 10 random seeds to ensure robustness.

%
\begin{figure}[t!]
  \centering
  \includegraphics[width=\linewidth]{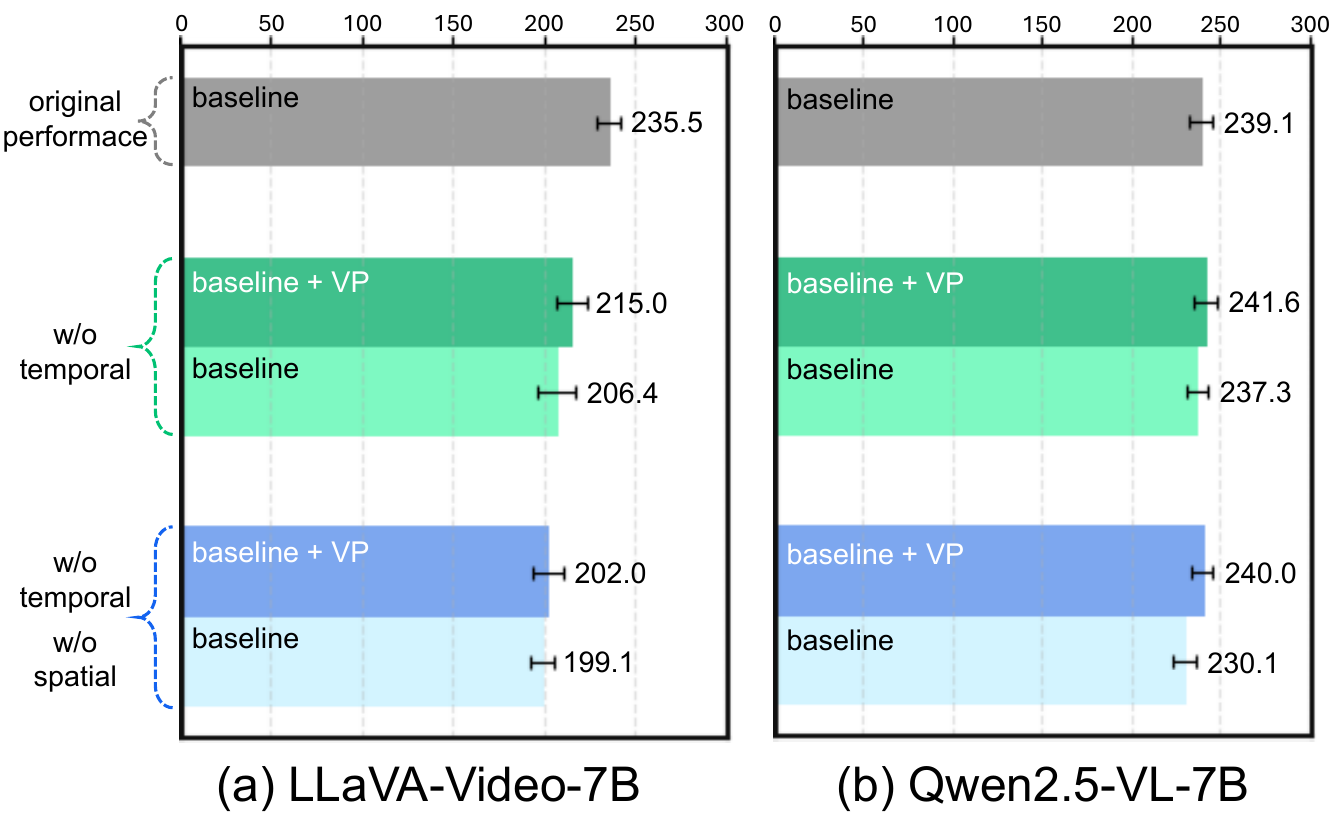}
  \vspace{-18 pt}
  \caption{
   \textbf{Effect of VP under Position Degradation.} Across both baselines, frame-index VP consistently improve accuracy under temporal-only and full-collapse settings, indicating added temporal cues and robustness to degraded positional signals.
   }
  \label{fig:fig_3}
\vspace{-10 pt}
\end{figure}
%

\subsection{Positional Embedding Degradation}

Recent Video-LLMs concatenate text and visual tokens into a single sequence and use rotary position embeddings (RoPE) to encode relative positions. Under this mechanism, frame order is reflected in the position indices assigned to visual tokens. In this section, as shown in \cref{fig:fig_2}(a), we manipulate these indices~\cite{vrope} to weaken temporal cues in the model and examine whether a VP can improve temporal understanding. In addition to LLaVA-Video (standard RoPE), we evaluate Qwen2.5-VL (M-RoPE)~\cite{qwen2vl} to test whether the observations persist under a different positional scheme.

\noindent \textbf{Temporal-only Degradation.}
In temporal-only degradation, we reassign position indices so that visual tokens from different frames share the same index range. This removes positional signals that encode frame order while preserving within-frame spatial order, allowing us to test whether VP restores frame-order awareness.

\noindent \textbf{Full Collapse.}
Full collapse further extends temporal-only degradation by additionally removing spatial information. All visual tokens share a single position index, eliminating both temporal and spatial cues. This configuration acts as a controlled extreme case, designed to evaluate VP performance when all positional guidance is entirely removed.

\noindent \textbf{Findings.}
As shown in \cref{fig:fig_3}, adding VP consistently improves accuracy across both baselines under positional degradation. The gains range from 2.9 to 9.9 points under degraded settings, with similar standard deviations across seeds. These results indicate that VP supplies additional temporal cues and serves as an effective mechanism for supporting the temporal understanding of VideoLLMs.

\subsection{Frame-level Referencing in VideoLLMs}
\label{sec:sec_3_2}

\subsubsection{Frame Number as a Dictionary Key}
We investigate whether VideoLLMs, which process inputs as a flattened sequence of frame tokens, can still treat the \textit{N-th frame} as a distinct unit. We further examine whether adding a VP strengthens this frame-level awareness and enables dictionary-like scene indexing. To assess this, we designed an experiment, illustrated in ~\cref{fig:fig_2}(b), which utilizes unique videos from the VideoMME temporal reasoning dataset. Each video was uniformly sampled into 8 / 16 / 32 / 64 frames, with a synthetic panda image randomly inserted at a specific frame. We then evaluated the model's ability to perform: 1) Lookup and 2) Reverse Lookup, using the frame number as the \textit{key}.

\vspace{1pt}
\noindent \textbf{Lookup: Explain frame $k$.} 
To evaluate lookup capability, we used the frame number $k$ of the inserted panda image as a key, querying the model to describe the content of that specific frame. We measured accuracy by verifying the presence of the word `panda' in the generated response. The baseline model (without VP) showed poor accuracy, averaging 12.43\% and plummeting to just 4.09\% as the frame count increased to 64. This indicates the baseline struggles to grasp the ordinal concept of the \textit{N-th frame}. Conversely, applying VP yields a substantial 4–5× gain in average performance, indicating that it helps the model more accurately pinpoint information at specific moments.

\noindent \textbf{Reverse Lookup: Where is panda?} 
To function as a dictionary, the model must support both key-to-value (lookup) and value-to-key (reverse lookup) retrieval. Accordingly, we asked the model to identify the exact frame index containing a panda. Although this resembles a temporal grounding task, our evaluation is more strict, as it requires pinpointing a single frame rather than a range. The baseline model's performance declined significantly with increased frame counts, reaching only 3.51\% accuracy at 64 frames. In comparison, the VP-augmented model maintained high accuracy, achieving perfect results at specific positions. This demonstrates that the VP mechanism effectively enables precise reverse lookup based on visual input.

\subsubsection{Positional bias of Visual Prompting}

We observed a performance gap depending on the placement of the VP. As shown in Table~\ref{tab:tab_1}, positioning the frame number in the bottom-left (BL) or bottom-right (BR) corners consistently led to higher accuracy in both lookup and reverse lookup tasks compared to the top-left (TL) and top-right (TR) positions. This effect was especially pronounced in reverse lookup: BL and BR achieved 100\% accuracy regardless of frame count, while TL and TR yielded significantly lower averages of 60.19\% and 78.70\%, respectively.
These results suggest that the model does not attend uniformly across the frame, but rather places greater emphasis on cues in the lower region. This tendency likely reflects biases in the training data, where subtitles, timestamps, or watermarks frequently appear at the bottom of web-based videos, influencing the model’s vision-language alignment.
To better understand this behaviour, we inspected errors in the TL setting for reverse lookup. In almost all such cases, the model predicted the frame immediately after the correct one. We hypothesize that this off-by-one shift arises because frame features are concatenated into a single token sequence without explicit boundaries. In this setting, the top-left number prompt can be confused with the tokens of the following frame rather than those of the current one.

\begingroup
\renewcommand{\arraystretch}{1.0}
\begin{table}[t]
  \centering
  \caption{
    \textbf{Results on lookup and reverse-lookup tasks.} The table compares performance across VP placements, with and without VP, for mapping frame numbers to scenes (lookup) and scenes back to frame numbers (reverse lookup).
    }
  \vspace{-8pt}
  \resizebox{\columnwidth}{!}{
  \begin{tabular}{lccccc}
    \toprule
    \multicolumn{6}{c}{\textbf{(a) Look-up}} \\
    \midrule
    Position & 8 frames & 16 frames & 32 frames & 64 frames & average \\
    \midrule
    --  & 28.07 & 11.70 & 5.85  & 4.09  & 12.43 \\
    TL  & 67.25 & 60.23 & 55.56 & 39.18 & 55.56  \\
    TR  & 56.73 & 53.80 & 51.46 & 31.58 & 48.39 \\
    BL  & \textbf{77.19} & \textbf{64.91} & 63.16 & \textbf{53.22} & \textbf{64.62}  \\
    BR  & 76.02 & 64.33 & \textbf{64.33} & 51.46 & 64.04  \\
    \midrule
    \multicolumn{6}{c}{\textbf{(b) Reverse Look-up}} \\
    \midrule
    Position & 8 frames & 16 frames & 32 frames & 64 frames & average \\
    \midrule
    --  & 45.03 & 16.96 & 8.77  & 3.51  & 18.57 \\
    TL  & 79.53 & 63.57 & 53.80 & 43.86 & 60.19   \\
    TR  & 73.10 & 83.04 & 85.38 & 70.76 & 78.70   \\
    BL  & \textbf{100.0} & \textbf{100.0} & \textbf{100.0} & \textbf{100.0} & \textbf{100.0} \\
    BR  & \textbf{100.0} & \textbf{100.0} & \textbf{100.0} & \textbf{100.0} & \textbf{100.0} \\
    \bottomrule
  \end{tabular}
  }
  \label{tab:tab_1}
  \vspace{-10pt}
\end{table}
\endgroup

%
\begin{figure*}[t!]
  \centering
  \includegraphics[width=\linewidth]{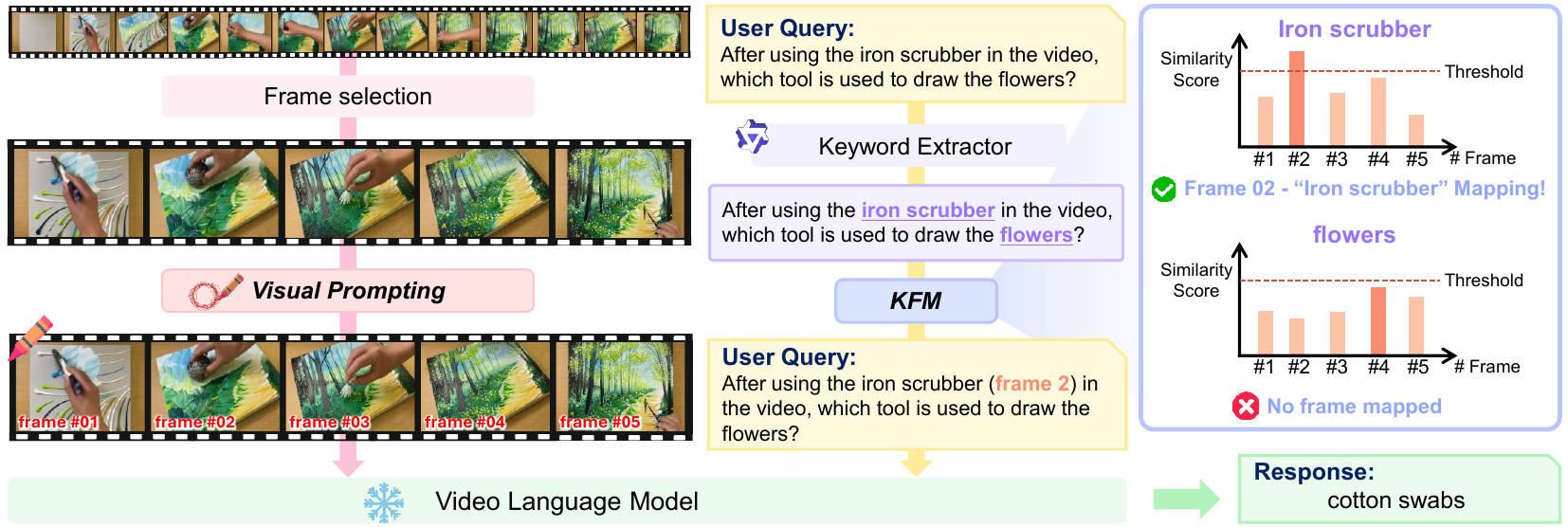}
  \vspace{-18 pt}
  \caption{\textbf{Overall pipeline of \ourmodel.}  Given a sequence of video frames, we first apply visual prompting by overlaying frame-number VPs on each frame. For a user query, the system extracts key textual concepts and performs Keyword–Frame Mapping (KFM) to identify which frames best correspond to each keyword. The query is then rewritten to explicitly include the mapped frame indices, aligning textual cues with the numbered frames. Feeding this aligned query–frame pair into the VLM enables more accurate temporal understanding.
  }
  \label{fig:fig_4}
\vspace{-10 pt}
\end{figure*}
%

\subsection{Attention-based Analysis}

Previous studies have consistently reported that increased attention from text tokens to image tokens is associated with improved video understanding performance~\cite{attention_cite1, attention_cite2, attention_cite3}. To investigate how VP influence this behavior, we measured the average attention allocated to image tokens at each layer, comparing models with and without VP.

\noindent \textbf{Layer-wise Attention Shift.} 
As shown in Fig.~\ref{fig:fig_2}(c), the presence of VP leads to higher attention on image tokens across all layers, with an average relative increase of 11.65\% ($\pm$0.18). The overall attention also increases from 0.081 ($\pm$0.05) in the noVP setting to 0.089 ($\pm$0.05) with VP. Importantly, this change occurs without modifying the number of image tokens. This indicates that VP adds no new inputs and instead shifts attention to existing content.

\noindent \textbf{Localized Effects Across Depth.} 
Although the overall increase is moderate, certain layers exhibit notably larger gains. In particular, attention rises are concentrated in mid-to-late layers such as layer 4–6, 11–14, and after 21. This suggests that VP has a stronger effect during stages related to spatiotemporal integration and vision-language alignment, rather than in the early stages of visual encoding. These results complement our earlier findings on frame lookup and reverse lookup tasks, indicating that VP helps the model localize and attend to temporally relevant frames by guiding attention toward image tokens near the specified cue, thereby enhancing temporal reasoning performance.

\section{Method}
\label{sec:method}
\subsection{Preliminaries}

We approach video understanding using a VideoLLM that jointly processes visual and textual inputs to generate natural language outputs. Given an input video $\mathbf{V}$, we first apply a frame sampling strategy $\mathcal{S}$ to extract a subset of $N$ representative frames: 
\begin{equation}
\small
\mathbf{F} = \mathcal{S}(\mathbf{V}) = \{ \mathbf{I}_1, \ldots, \mathbf{I}_N \}.
\end{equation}
Here, $\mathbf{I}_i$ denotes the $i$-th sampled frame, where $i = 1, \ldots, N$.
Each sampled frame $\mathbf{I}_i$ is encoded into $P$ visual tokens using a pretrained vision encoder $\phi(\cdot)$:  
$\phi(\mathbf{I}_i) = [\mathbf{v}_{i,1}, \ldots, \mathbf{v}_{i,P}]$, where $\mathbf{v}_{i,p} \in \mathbb{R}^d$.
Stacking tokens over $N$ frames yields the full visual-token sequence:
\begin{equation}
\small
\mathbf{Z}_{\text{vis}}=\big[\phi(\mathbf{I}_1)\,\Vert\,\cdots\,\Vert\,\phi(\mathbf{I}_N)\big] \in \mathbb{R}^{(NP) \times d}.
\end{equation}
The text prompt $\mathbf{p}$ is tokenized by a language encoder $\psi(\cdot)$ into text embeddings $ \mathbf{Z}_{\text{text}}=\big[\mathbf{t}_1,\ldots,\mathbf{t}_L\big] $ with $ \mathbf{t}_\ell\in\mathbb{R}^{d} $.
The model then processes the concatenated multimodal sequence and produces a natural language output:
\begin{equation}
\small
\hat{y}=\text{VideoLLM}\!\left(\big[\mathbf{Z}_{\text{vis}}\,\Vert\,\mathbf{Z}_{\text{text}}\big]\right).
\end{equation}

\subsection{Sequential Visual Prompting}
\label{sec:sec_4_2_vp}
We insert a number-based VP directly into the pixel space of each video frame. For the $i$-th sampled frame, the VP-injected image is denoted as:
\begin{equation}
\small
\hat{\mathbf{I}}_i = \text{InsertVP}(\mathbf{I}_i, i, \text{fontsize}),
\end{equation}
where \( \mathbf{I}_i \) is the original image and $i$ represents its position in the sampled sequence. Based on the positional bias observed in~\cref{sec:sec_3_2}, we place the VP consistently at the bottom-left corner of the frame.

To adapt the prompt size to videos with varying resolutions, the font size is defined based on the dimensions of each frame $\mathbf{I}_i$ as
$\text{fontsize} = \min(\text{width}(\mathbf{I}_i), \, \text{height}(\mathbf{I}_i)) / s$,
where $s$ is a hyperparameter controlling the relative size of the prompt, and the division result is floored to ensure an integer font size. Detailed parameter settings are provided in the supplementary material.

\begin{table*}[t]
\centering
\caption{
\textbf{Main results.} Results on the temporal understanding subsets of the TempCompass~\cite{tempcompass}, MVBench~\cite{mvbench}, VideoMME~\cite{videomme}, and LongVideoBench~\cite{longvideobench} datasets. 
Our VP-based method consistently improves performance across all datasets, and even when the input is limited to 20\% of the frames, it achieves accuracy comparable to the 1\,FPS dense-frame setting in some cases.
}

\resizebox{\linewidth}{!}{%
\begin{tabular}{p{0.2cm} l!{\vrule}ccc!{\vrule}ccc!{\vrule}c!{\vrule}ccccc}
\hline
& \multirow{2}{*}{Models} 
& \multicolumn{3}{c|}{TempCompass}
& \multicolumn{3}{c|}{MVBench}
& \multicolumn{1}{c|}{VideoMME}
& \multicolumn{4}{c}{LongVideoBench}
\\
& & EO & AC & total & AS & ST & total & TR & E3E & O3O & SSS & total \\
\hline

\multirow{10}{*}{\centering\rotatebox{90}{\textbf{64 Frames (1 FPS)}}}
& \multicolumn{12}{>{\columncolor{gray!20}}l}{\textit{VideoLLM Baselines}}\\
& GPT4.1~\cite{gpt4_1} & 66.69 & 74.72 & 70.74 & 75.00 & 92.50 & 83.75 & $-$ & $-$ & $-$ & $-$ & $-$  \\ 
& Qwen2.5-VL-7B~\cite{qwen2vl} & 75.00 & 80.26 & 77.65 & 79.50 & 92.00 & 85.75 & 38.42 & 57.45 & 48.48 & 40.21 & 48.63 \\
& LLaVA-Video-7B~\cite{llavavideo} & 73.63 & 75.71 & 74.68 & 74.00 & 91.00 & 82.50 & 47.46 & 69.15 & 62.12 & 40.21 & 56.42 \\
& LLaVA-OneVision-7B~\cite{llavaonevision} & 71.24 & 71.66 & 71.45 & 73.00 & 92.00 & 82.50 & 43.50 & 67.02 & 56.06 & 36.08 & 52.53 \\
& \multicolumn{12}{>{\columncolor{gray!20}}l}{\textit{VideoLLM w/ \emph{VP}}}\\

& Ours~\textsubscript{\textit{GPT4.1}} & 
76.66 & 84.02 & 80.37~\textsubscript{\textcolor{red}{$\blacktriangle$}} & 
74.50 & 93.00 & 83.75~\textsubscript{\textcolor{gray}{\rule{0.6em}{0.6ex}}}
& $-$ & $-$ & $-$ & $-$ & $-$ \\

& Ours~\textsubscript{\textit{QwenVL}} & 
76.95 & 80.47 & 78.72~\textsubscript{\textcolor{red}{$\blacktriangle$}} & 
79.0 & 96.12 & 87.68~\textsubscript{\textcolor{red}{$\blacktriangle$}} & 
41.25~\textsubscript{\textcolor{red}{$\blacktriangle$}} & 
60.64 & 50.00 & 38.14 & 49.42~\textsubscript{\textcolor{red}{$\blacktriangle$}} \\

& Ours~\textsubscript{\textit{LLaVA-Video}}&
73.99 & 77.7 & 75.86~\textsubscript{\textcolor{red}{$\blacktriangle$}} & 
71.50 & 91.00 & 81.25~\textsubscript{\textcolor{blue}{$\blacktriangledown$}} &
48.02~\textsubscript{\textcolor{red}{$\blacktriangle$}} & 
73.40 & 57.58 & 39.18 & 56.42~\textcolor{gray}{\rule{0.45em}{0.4ex}} \\

& Ours~\textsubscript{\textit{LLaVA-OneVision}}& 
73.48 & 72.16 & 72.82~\textsubscript{\textcolor{red}{$\blacktriangle$}}& 
70.00 & 90.50 & 80.25~\textsubscript{\textcolor{blue}{$\blacktriangledown$}}& 
42.94~\textsubscript{\textcolor{blue}{$\blacktriangledown$}} & 
69.15 & 54.55 & 40.21 & 54.47~\textsubscript{\textcolor{red}{$\blacktriangle$}}\\
\hline
\hline

\multirow{10}{*}{\centering\rotatebox{90}{\textbf{20\% Frames}}}
& \multicolumn{12}{>{\columncolor{gray!20}}l}{\textit{VideoLLM Baselines}}\\

& GPT4.1~\cite{gpt4_1} & 64.45&79.90&72.24&68.50&92.00&80.25& $-$ & $-$ & $-$ & $-$ & $-$ \\
& Qwen2.5-VL-7B~\cite{qwen2vl} & 72.18 & 73.86 & 73.03 & 73.00 & 90.00 & 81.50 & 42.94 & 52.13 & 42.42 & 34.02 & 42.80  \\
& LLaVA-Video-7B~\cite{llavavideo} & 66.11 & 65.55 & 65.53 & 64.50 & 89.50 & 77.00 & 44.07 & 63.83 & 48.48 & 32.99 & 48.25 \\

& LLaVA-OneVision-7B~\cite{llavaonevision}  & 70.81 & 67.54 & 69.16 & 69.00 & 89.50 & 79.25 & 43.50 & 63.83 & 46.97 & 34.02 & 48.25 \\
& \multicolumn{12}{>{\columncolor{gray!20}}l}{\textit{VideoLLM w/ \emph{VP}}}\\

& Ours~\textsubscript{\textit{GPT4.1}} & 
75.51 & 83.52 & 79.55~\textsubscript{\textcolor{red}{$\blacktriangle$}} &
67.00 & 94.00 & 80.50~\textsubscript{\textcolor{red}{$\blacktriangle$}} & 
$-$ & $-$ & $-$ & $-$ & $-$ \\

& Ours~\textsubscript{\textit{QwenVL}} & 
75.22 & 79.97 & 77.61~\textsubscript{\textcolor{red}{$\blacktriangle$}} & 
73.00 & 91.00 & 82.00~\textsubscript{\textcolor{red}{$\blacktriangle$}} & 
44.07~\textsubscript{\textcolor{red}{$\blacktriangle$}} & 
54.26 & 45.45 & 32.99 & 43.97~\textsubscript{\textcolor{red}{$\blacktriangle$}} \\

& Ours~\textsubscript{\textit{LLaVA-Video}}& 
72.33 & 72.09 & 72.21~\textsubscript{\textcolor{red}{$\blacktriangle$}} & 
68.50 & 91.00 & 79.75~\textsubscript{\textcolor{red}{$\blacktriangle$}} &
50.28~\textsubscript{\textcolor{red}{$\blacktriangle$}} & 
62.77 & 57.58 & 32.99 & 50.19~\textsubscript{\textcolor{red}{$\blacktriangle$}} \\
& Ours~\textsubscript{\textit{LLaVA-OneVision}}&
71.97 & 67.54 & 69.73~\textsubscript{\textcolor{red}{$\blacktriangle$}} &
67.50 & 90.50 & 79.00~\textsubscript{\textcolor{blue}{$\blacktriangledown$}}& 
44.63~\textsubscript{\textcolor{red}{$\blacktriangle$}}& 
64.89 & 51.52 & 41.24 & 52.53~\textsubscript{\textcolor{red}{$\blacktriangle$}}\\
\bottomrule
\end{tabular}%
}
\label{tab:tab2_main_results}
\end{table*}

\subsection{Keyword-Frame Mapping}
\label{sec:sec_4_3_mapping}
In \cref{sec:sec_3_2}, we observed that when provided with a frame index as a VP, the VideoLLM can utilize this index as a dictionary key, effectively associating it with specific visual content. 
Based on this observation, we propose a mapping-based prompting method, referred to as Keyword-Frame Mapping (KFM), which explicitly guides the model toward frames most relevant to the user query.

To implement this approach, we extract keywords from the user query $q$ that point to important visual content using an off-the-shelf LLM-based extractor:
\begin{equation}
\small
\mathbf{W} = \text{KeywordExtract}(q) = \{w_1, w_2, \ldots, w_m\}.
\end{equation}
We then compute the similarity between each keyword and the frames sampled from the video using a vision-language model such as CLIP~\cite{clip}. Let $\text{CLIP}_\text{img}(\hat{\mathbf{I}}_i)$ and $\text{CLIP}_\text{text}(w_j)$ denote the embeddings of frame $\hat{\mathbf{I}}_i$ and keyword $w_j$ , respectively. Their cosine similarity is computed as:
\begin{equation}
\small
\text{sim}(\hat{\mathbf{I}}_i, w_j) = \cos\!\left( \text{CLIP}_\text{img}(\hat{\mathbf{I}}_i), \text{CLIP}_\text{text}(w_j) \right).
\end{equation}
For each keyword $w_j$, we select the frame $\hat{\mathbf{I}}_{k^*}$ with the highest similarity:
\begin{equation}
\small
k^* = \arg\max_{i} \, \text{sim}(\hat{\mathbf{I}}_i, w_j).
\end{equation}
If the maximum similarity exceeds a predefined threshold $\tau$, the corresponding frame index $k^*$ is considered semantically relevant and selected for mapping into the prompt.
This condition is formally expressed as:
\begin{equation}
\small
\text{if } \text{sim}(\hat{\mathbf{I}}_{k^*}, w_j) \geq \tau \text{, then map } w_j \rightarrow k^*.
\end{equation}
Otherwise, no mapping is performed. This avoids assigning a keyword to a frame that lacks semantic relevance. For example, as shown in \cref{fig:fig_4}, the keyword \textit{flowers} has the highest similarity to frame 4, but the score remains below the threshold, indicating no match among sampled frames.

After applying this process to all keywords $w_j \in \mathbf{W}$, the original text embedding $\mathbf{Z}_{\text{text}}$ is transformed into an augmented version $\hat{\mathbf{Z}}_{\text{text}}$ by inserting the mapped frame indices:
\begin{equation}
\small
\hat{\mathbf{Z}}_{\text{text}} = \text{InsertIndex}(\mathbf{Z}_{\text{text}}, \{k^*_j\}_{j=1}^m),
\end{equation}
where $\{k^*_j\}$ denotes the set of frame indices mapped from keywords whose similarity exceeds the threshold.

This process guides the model to focus on specific frames during inference by explicitly linking keyword concepts to frame indices in the textual prompt, thereby providing temporal grounding based on keyword-guided semantic alignment between the textual query and visual content.

\section{Experiments}
\label{sec:experiments}

%
\begin{figure*}[t!]
  \centering
  \includegraphics[width=\linewidth]{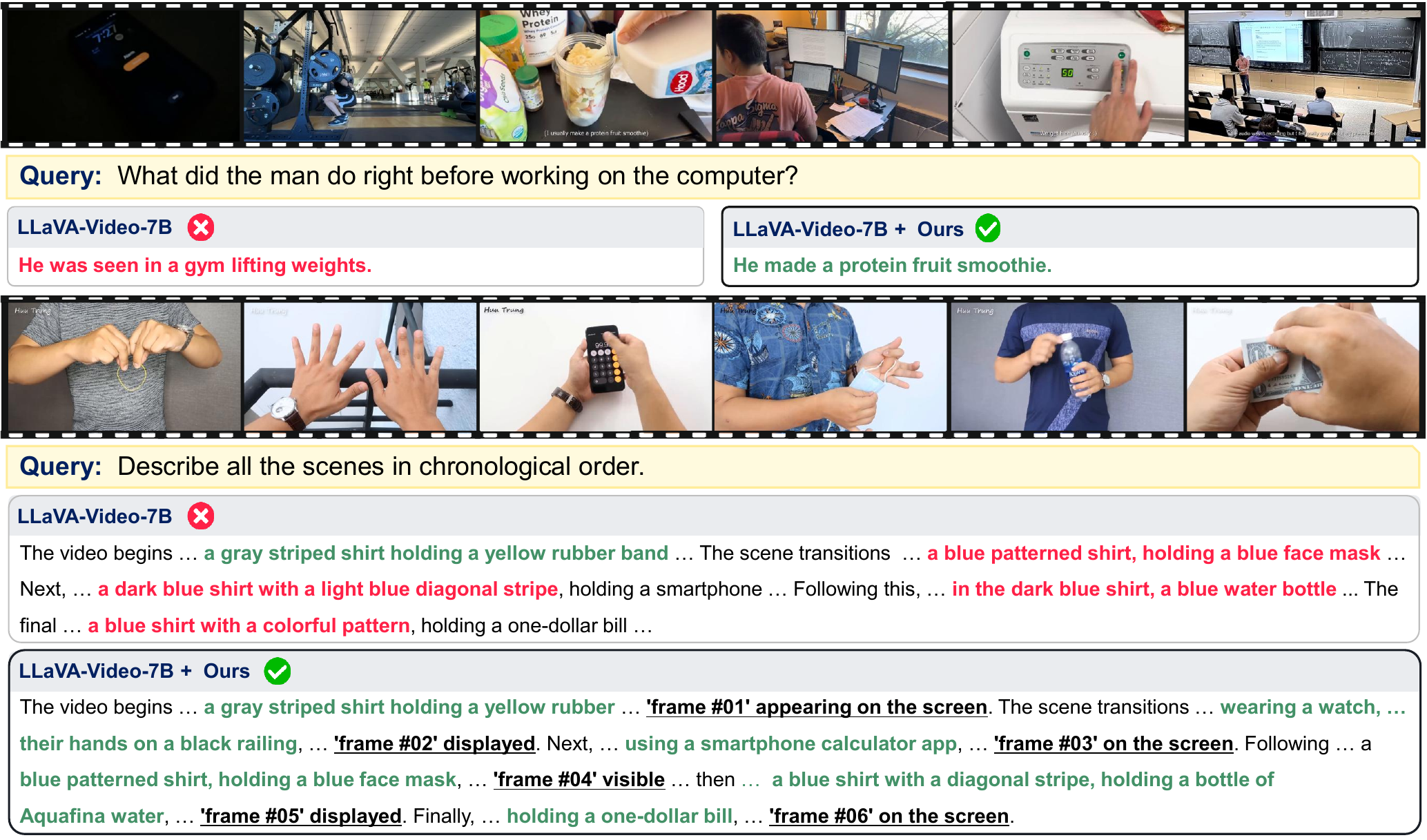}
  \vspace{-18 pt}
  \caption{\textbf{Qualitative Results on Open-Ended Temporal Questions.}
  Compared with the baseline LLaVA-Video-7B, \ourmodel{} uses explicit frame indices and keyword–frame mapping to better capture temporal order and thus produce more accurate answers to open-ended temporal queries. In particular, it often explicitly mentions the injected frame indices (\eg, ``frame \#01'', ``frame \#02'') in its free-form responses, indicating that its temporal reasoning is grounded on the visual prompts.}
  \label{fig:fig_5}
\vspace{-10 pt}
\end{figure*}
%

\begin{figure}[t!]
  \centering
  \includegraphics[width=\linewidth]{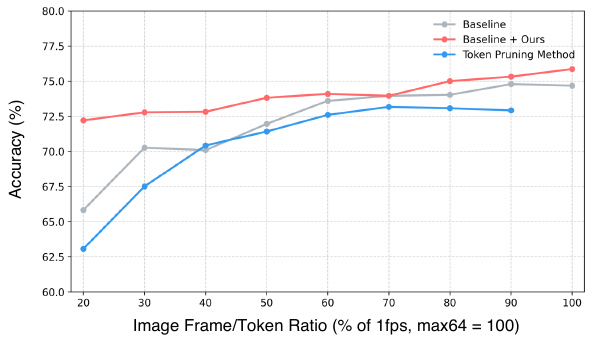}
  \vspace{-18 pt}
  \caption{
  \textbf{Performance Comparison under Varying Frame Sampling Ratios.} Accuracy on the TempCompass~\cite{tempcompass} dataset under different frame sampling ratios for our method and the baseline~\cite{llavavideo}, and under varying token reduction levels for the token pruning method~\cite{framefusion}. Our method with visual prompting consistently outperforms both baselines and maintains high accuracy even with fewer frames.
}
  \label{fig:fig_6}
\vspace{-18 pt}
\end{figure}

\subsection{Experimental Settings}
\label{sec:5_1}
\noindent \textbf{Datasets.} To evaluate the model's temporal understanding across diverse video lengths, content types, and question formats, we selected four benchmark datasets: TempCompass~\cite{tempcompass}, MVBench~\cite{mvbench}, VideoMME~\cite{videomme}, and LongVideoBench~\cite{longvideobench}. From each dataset, we sampled only the categories directly related to temporal understanding and excluded subtitles. Specifically, we used Event Order (EO) and Attribute Change (AC) from TempCompass~\cite{tempcompass}; Action Sequence (AS) and Scene Transition (ST) from MVBench~\cite{mvbench}; Temporal Reasoning (TR) from VideoMME~\cite{videomme}; and Event Before/After Event (E3E), Object Before/After Object (O3O), and Sequence of Scenes (SSS) from LongVideoBench~\cite{longvideobench}. 
Detailed information about the datasets is provided in the supplementary material, along with descriptions of key hyperparameters such as the font size controller $s$ and the KFM threshold $\tau$.

\noindent \textbf{Baselines and Metric.}
We evaluate five multimodal LLM baselines: GPT-4.1, Qwen2.5-VL-7B~\cite{qwen2vl}, LLaVA-Video-7B~\cite{llavavideo}, and LLaVA-OneVision-7B~\cite{llavaonevision}. We run each model via official APIs or publicly released checkpoints. Evaluation is performed using the Visual Question Answering (VQA) metric. All experiments are conducted on NVIDIA RTX A6000 and A100 GPUs.

\subsection{Evaluation Results}

To evaluate the effectiveness and generalizability of our method, we conduct both quantitative and qualitative experiments across a variety of benchmark datasets and baseline models. By applying our approach to multiple settings, we aim to assess its robustness and its impact on temporal reasoning performance under diverse conditions.

\noindent \textbf{Quantitative Results.} 
\Cref{tab:tab2_main_results} presents quantitative results under two different settings designed to simulate varying levels of temporal continuity. 
The \textit{64 Frames} setting corresponds to uniformly sampling up to 64 frames per video at 1 fps, resulting in relatively strong temporal coherence across frames. In contrast, the \textit{20\% Frames} setting involves uniformly sampling only 20\% of the frames from the same pool, leading to sparser and more temporally disjoint inputs. To ensure a minimum level of temporal context, we enforce a lower bound of three frames per video.
Across all benchmarks and datasets, our method consistently improves performance under the 20\% setting. Notably, on the VideoMME dataset, our method achieves even higher performance under the sparse 20\% condition compared to the dense 64-frame baseline without visual prompting. These results demonstrate the general effectiveness of our approach in enhancing temporal understanding, particularly when temporal cues are limited.
Additional results on non-temporal categories and keyframe selection for long videos are provided in the supplementary material.

\noindent \textbf{Efficient Performance across Sampling.}
In VideoLLMs, efficiency is commonly improved through frame sampling or token-level techniques such as token merging and pruning. To evaluate the trade-off between computational cost and performance, we compare our method not only with a strong baseline~\cite{llavavideo}, but also with FrameFusion~\cite{framefusion}, a state-of-the-art approach that improves efficiency through token reduction. The results are summarized in \cref{fig:fig_6}.
Our method consistently outperforms the baseline across a wide range of sampling ratios, from 20\% to 90\%, often achieving comparable or even superior accuracy to the 100\% frame setting, despite using significantly fewer frames. In contrast, token-based methods show slightly lower accuracy than frame-sampled baselines, suggesting that they may be less suitable for temporal understanding tasks. This may be due to the fact that pruning or merging similar tokens can reduce the continuity of scene-level and temporal information, which is essential for reasoning over time.

\noindent \textbf{Qualitative Results.}
We present qualitative results on open-ended queries that are difficult to capture with standard multiple-choice VQA metrics.
Free-form responses are useful because they reveal not only whether the answer is correct, but also how the model reasons about the video~\cite{openended}.
By injecting frame indices as visual cues and mapping keywords to frames, our model achieves more accurate answers to temporal questions.
In the top of~\cref{fig:fig_5}, it aligns the keyword \textit{working on the computer} with the fourth frame and then correctly identifies the third frame as the immediately preceding event, answering the question about what happened right before the computer scene.
In the bottom of~\cref{fig:fig_5}, the model describes the scenes in temporal order and explicitly mentions the injected frame indices (the underlined parts), demonstrating the use of visual prompts to recover the correct event ordering.
Overall, these examples show that \ourmodel{} is effective for answering open-ended questions that require temporal ordering.

\subsection{Ablation Study}

\noindent \textbf{Effect of VP and KFM.}
As shown in \cref{tab:tab3_ablation}(a), both VP and KFM individually improve performance, but their combination yields the best results. VP alone provides clear gains by enhancing frame-level understanding, while KFM helps guide attention to relevant frames. As demonstrated in \cref{sec:sec_3_2}, providing frame indices via VP strengthens the effect of mapping by grounding it in the visual timeline, allowing the model to better interpret index-based references.

\begingroup 
\begin{table}[t!]
\centering
\caption{\textbf{Ablation Results.} 
Part (a) shows effect of VP and KFM, and part
(b) shows effect of different keyword extractors and mapping backbones. Results are reported on VideoMME (V-MME)~\cite{videomme} and LongVideoBench~\cite{longvideobench}.}
\vspace{-8pt}
\small

\begin{tabular}{c cc c cccc}
\hlineB{2.5}
\multicolumn{8}{l}{(a) VP and KFM}\\
\hlineB{1.5}
{\footnotesize Exp.} & \multicolumn{1}{c}{\multirow{2}{*}{VP}} & 
\multicolumn{1}{c}{\multirow{2}{*}{KFM}} &
\multicolumn{1}{c}{V-MME} &
\multicolumn{4}{c}{LongVideoBench} \\
\# & & &
TR & E3E & O3O & SSS & total \\
\hlineB{1.5}
1 &  &  & 44.1 & 63.8 & 48.5 & 33.0 & 48.3 \\
\rowcolor{gray!5}
2 & \checkmark &  & 49.2 & 62.8 & 54.6 & 32.0 & 49.0 \\
3 &  & \checkmark & 45.2 & 62.8 & 51.5 & 37.1 & 50.2 \\
\hlineB{1.5}
\rowcolor{gray!10}
4 & \checkmark & \checkmark & 50.3 & 62.8 & 57.6 & 33.0 & 50.2 \\
\hlineB{2.5}
\end{tabular}

\vspace{5pt}

\begin{tabular}{l c c cc c}
\hlineB{2.5}
\multicolumn{6}{l}{(b) Keyword Extractor and Mapping Backbone}\\
\hlineB{1.5}
\multirow{2}{*}{} &
\multicolumn{1}{c}{V-MME} &
\multicolumn{4}{c}{LongVideoBench} \\
\multicolumn{1}{l}{} & TR & E3E & O3O & SSS & total \\
\hline
\rowcolor{gray!10}
\textit{Keyword Extractor} &  &  &  &  &  \\
GPT-4o & 50.3 & 63.8 & 54.6 & 32.0 & 49.4 \\
\rowcolor{gray!5}
Qwen-1.5B & 50.3 & 64.9 & 54.6 & 32.0 & 49.8 \\
Qwen-3B & 50.3 & 64.9 & 54.6 & 32.0 & 49.8 \\
\rowcolor{gray!5}
Qwen-7B (Ours) & 50.3 & 62.8 & 57.6 & 33.0 & 50.2 \\
\hlineB{1.5}
\rowcolor{gray!10}
\textit{Mapping Backbone} &  &  &  &  &  \\
CLIP-B/32 & 42.9 & 61.7 & 53.0 & 32.0 & 48.3 \\
\rowcolor{gray!5}
CLIP-L/14 (Ours) & 50.3 & 62.8 & 57.6 & 33.0 & 50.2 \\
\hlineB{2.5}
\end{tabular}

\label{tab:tab3_ablation}
\end{table}
\endgroup

\noindent\textbf{Keyword Extractor and Mapping Backbone.}  
\cref{tab:tab3_ablation}(b) shows the performance variation using different keyword extractors such as GPT-4o, Qwen-1.5B, Qwen-3B, and Qwen-7B, along with two vision-language mapping backbones, CLIP-B/32 and CLIP-L/14. For keyword extraction, performance remains consistently strong across models except for the smallest variant. Regarding the mapping backbone, CLIP-B and CLIP-L perform similarly overall, as long as the similarity threshold is tuned appropriately for each to account for differences in embedding scale.

\begin{figure}[t!]
  \centering
  \includegraphics[width=\linewidth]{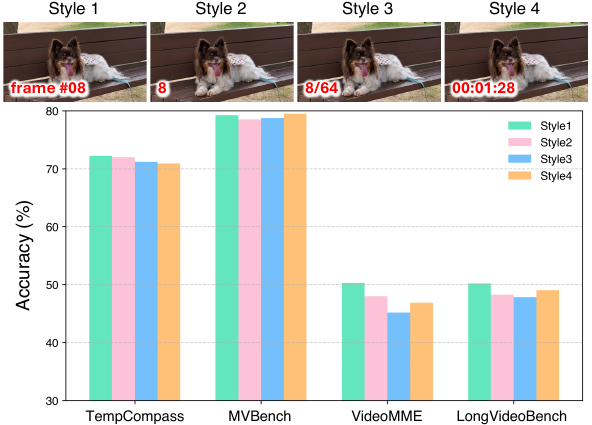}
  \vspace{-18 pt}
  \caption{
  \textbf{Performance comparison across various VP styles.} Styles 1 to 4 differ in how sequential cues are visually represented, and their impact is evaluated across four video understanding benchmarks.
  }
  \label{fig:fig_7}
\vspace{-10 pt}
\end{figure}

\noindent\textbf{Design Choices for Sequential VP.}  
\cref{fig:fig_7} compares four VP styles: Styles 1–3 use different formats for displaying the frame number, while Style 4 encodes the timestamp instead. Among them, Style 1 (ours) consistently achieves the best performance across all four benchmarks. We observed that simple red text can become difficult to recognize depending on the dataset and visual background. To address this, we optionally include a bordered background to improve visibility, which is treated as a hyperparameter and further detailed in the supplementary material.

%
%

\section{Conclusion}
\label{sec:conclusion}

We introduce \ourmodel, a simple yet effective method to enhance temporal understanding in VideoLLMs through the use of sequential VP and KFM. By inserting frame numbers into the visual input, our method provides explicit temporal cues for temporally sparse video inputs. Combined with text-to-frame alignment, this guidance improves performance across diverse video reasoning tasks and sampling conditions with minimal overhead. Experiments on multiple benchmarks show the generalizability and efficiency of our approach. Limitations and potential future directions are discussed in detail in the supplementary material.
\section*{Acknowledgment}
This work was supported in part by the IITP RS-2024-00457882 (AI Research Hub Project), IITP 2020-II201361, NRF RS-2024-00345806, NRF RS-2023-002620, and RQT-25-120390. Affiliations: Department of Artificial Intelligence (Y.L, D.J, Y.K, S.J.H), Department of Computer Science (S.K).
{
    \small
    \bibliographystyle{ieeenat_fullname}
    \bibliography{main}
}
\clearpage
\appendix
\onecolumn



\clearpage

\section{Analysis Details}

\subsection{Details of Positional Embedding Degradation}
We now describe how we manipulate rotary position embeddings under the temporal-only degradation and full-collapse settings.
\paragraph{Standard RoPE.}
Standard rotary position embeddings (RoPE) encode token order using a single 1D rotational embedding over the concatenated text--vision sequence, such that each visual token receives a unique position index along the sequence dimension. 
Let $L_t$ denote the number of text tokens, $N$ the number of visual tokens per frame, $k$ the (1-based) frame index, and $j$ the spatial token index within a frame.
In the original setting, the position index assigned to a visual token is
\[
\text{pos}(k,j) = L_t + (k-1)N + j,
\]
where the offset $(k-1)N$ encodes frame order and $j$ preserves within-frame spatial structure.

Under \textit{temporal-only degradation}, we remove the frame-dependent offset and assign the same index pattern to all frames:
\[
\text{pos}(k,j) \leftarrow L_t + j.
\]
This eliminates positional signals that encode frame order while preserving spatial ordering within each frame.
Under \textit{full collapse}, we further map all tokens to a single index,
\[
\text{pos}(k,j) \leftarrow L_t,
\]
removing both temporal and spatial positional distinctions.

\paragraph{M-RoPE.}
Multi-scale RoPE (M-RoPE) extends standard RoPE by decomposing positional information into multiple components, typically a temporal index and spatial indices, which are applied at different granularities to the Q/K representations.
Each visual token is associated with a triplet
\[
(t, h, w),
\]
where $t$ denotes the temporal index (frame) and $(h, w)$ encode spatial coordinates within the frame.

In the original M-RoPE configuration, temporal ordering is represented by variation in $t$ across frames, and $(h,w)$ preserve the spatial layout.
Under \textit{temporal-only degradation}, we fix the temporal component across frames while keeping spatial components unchanged:
\[
(t, h, w) \leftarrow (t_0, h, w),
\]
for some constant $t_0$, thereby removing temporal order but preserving spatial structure.
Under \textit{full collapse}, we map all tokens to a single triplet,
\[
(t, h, w) \leftarrow (t_0, h_0, w_0),
\]
for fixed $(t_0, h_0, w_0)$, eliminating both temporal and spatial positional cues in the embedding.

\paragraph{Relation to M-RoPE.}
This experiment also helps clarify the distinction between VP and M-RoPE. Although Qwen2.5 employs M-RoPE to encode temporal order within the embedding space, this mechanism operates implicitly through relative token positions. In contrast, VP introduces an explicit temporal signal in pixel space during visual encoding. As shown in Fig.~3(b) of the main paper, VP consistently improves performance even when positional information is degraded, indicating that its contribution goes beyond positional embeddings.


\subsection{Details of Frame-Level Referencing}

\paragraph{Settings.}
We construct a synthetic frame-referencing set from the temporal reasoning split of VideoMME by uniformly sampling $N \in {8,16,32,64}$ frames per video and overlaying a black-and-white panda image on exactly one randomly chosen frame.
Frame-index prompts (frame \#i) are inserted at a fixed corner (TL, TR, BL, or BR) with font size controlled as in Sec.~4.2 of the main paper (hyperparameter $s{=}12$), while all other visual content is kept unchanged.

\begingroup
\renewcommand{\arraystretch}{1.0}
\begin{table}[!ht]
  \centering
  \caption{
    Results showing the accuracy of reverse lookup using VP, measured with a tolerance of $\pm$1 frame.
    }
  \vspace{-8pt}
  \begin{tabular}{lccccc}
    \toprule
    Position & 8 frames & 16 frames & 32 frames & 64 frames & average \\
    \midrule
    --  & 69.01 & 35.67 & 17.54  & 7.60  & 32.46 \\
    TL  & \textbf{100.0} & 99.42 & \textbf{100.0} & 99.42 & 99.71 \\
    TR  & \textbf{100.0} & 99.42 & \textbf{100.0} & 97.66 & 99.27 \\
    BL  & \textbf{100.0} & \textbf{100.0} & \textbf{100.0} & \textbf{100.0} & \textbf{100.0} \\
    BR  & \textbf{100.0} & \textbf{100.0} & \textbf{100.0} & \textbf{100.0} & \textbf{100.0} \\
    \bottomrule
  \end{tabular}
  \label{tab:supple1_reverselookup}
  \vspace{-10pt}
\end{table}
\endgroup

\paragraph{Additional Positional Bias Results.} 
In Sec. 3.2 of the main paper, we demonstrated that inserting VP enables the model to recognize frame indices and use them as references for frame-level lookup and reverse lookup. In the reverse lookup experiments, we observed that placing the VP at the bottom-left (BL) or bottom-right (BR) corners led to perfect accuracy, whereas top-left (TL) and top-right (TR) placements resulted in reduced accuracy, with a notable tendency for the model to answer with adjacent frame numbers (either immediately before or after the target). \Cref{tab:supple1_reverselookup} shows the accuracy under a relaxed criterion allowing an error margin of $\pm$1 frame. Under this setting, even TL and TR positions achieved near-perfect accuracy across all frame counts.

\begin{table}[b!]
    \centering
    \caption{\textbf{Statistics of the analysis subsets.} For each dataset, we report the number of questions, number of videos, and video length statistics (in seconds) for the 100-example evaluation subsets used in our analysis.}
    \vspace{2pt}
    \begin{tabular}{l c c c r r r r}
        \hlineB{2.5}
        Dataset & Category & Questions & Videos & Min (s) & Max (s) & Mean (s) & Std (s) \\
        \hline
        \multirow{3}{*}{TempCompass} 
            & EO    & 55  & 45 & 3.04 & 27.20 & 14.59 & 6.99 \\
            & AC    & 45  & 39 & 3.00 & 24.68 & 10.46 & 5.19 \\
            & Total & 100 & 84 & 3.00 & 27.20 & 12.67 & 6.55 \\
        \hdashline
        \multirow{3}{*}{MVBench} 
            & AS    & 51  & 49 & 8.20 & 43.20 & 21.12 & 8.67 \\
            & ST    & 49  & 49 & 20.00 & 20.00 & 20.00 & 0.00 \\
            & Total & 100 & 98 & 8.20 & 43.20 & 20.56 & 6.16 \\
        \hdashline
        \multirow{1}{*}{VideoMME} 
            & TR    & 100 & 99 & 56.89 & 3540.14 & 1474.89 & 1078.71 \\
        \hdashline
        \multirow{4}{*}{LongVideoBench} 
            & E3E   & 31  & 28 & 11.50 & 2186.42 & 735.13 & 594.84 \\
            & O3O   & 30  & 24 & 7.98  & 2041.01 & 564.30 & 575.01 \\
            & SSS   & 39  & 29 & 8.57  & 1652.82 & 550.32 & 460.63 \\
            & Total & 100 & 81 & 7.98  & 2186.42 & 618.35 & 550.89 \\
        \hlineB{2.5}
    \end{tabular}
\label{tab:dataset_analysis}
\end{table}
\subsection{Details of Attention-Based Analysis}

To better understand how visual prompting (VP) influences the attention mechanism in VideoLLMs, we perform a layer-wise attention analysis using the prefilling stage of the language model. Specifically, we extract cross-modal attention values from text tokens (queries) to image tokens (keys) before any autoregressive decoding takes place. This allows us to isolate how much attention the model allocates to the visual input based solely on the prompt.

\paragraph{Settings.}
We use the same input video and prompt across conditions with and without VP, and compare the resulting attention distributions. The prompt consists of the full user query as used in each benchmark~\cite{tempcompass,mvbench,videomme,longvideobench}.
For each benchmark, we randomly sample 100 examples using seed 42 to ensure consistency across experiments.

\paragraph{Implementation.}
We modify the model's forward pass to cache the attention weights from all self-attention layers during the prefilling stage. The analysis is conducted on a LLaVA-Video~\cite{llavavideo} model with 32 transformer layers. For each layer, we average the attention values across all heads to obtain a single attention score per layer.

\paragraph{Computation.}
Let $A^{(l,h)} \in \mathbb{R}^{T \times S}$ denote the attention matrix at layer $l$ and head $h$, where $T$ is the number of text tokens and $S$ is the number of image tokens. We compute the mean attention over all query tokens and attention heads:
\begin{equation}
\bar{A}^{(l)} = \frac{1}{H} \sum_{h=1}^{H} \frac{1}{T} \sum_{t=1}^{T} \sum_{s=1}^{S} A_{t,s}^{(l,h)},
\end{equation}
where $H$ is the total number of attention heads. 
We then calculate the relative change in attention between the VP and no-VP settings at each layer, as reported in Fig. 2(c) of the main paper.

\paragraph{Query Selection.}
Following the convention in the prefilling stage of VideoLLMs, we use the final token of the input prompt (i.e., the last token of the user query) as the attention query. This aligns with how the model typically attends to visual inputs before generating an answer.

\paragraph{Interpretation.}
The results show that VP increases attention to image tokens without changing the number of image tokens themselves. The increase is especially prominent in mid-to-late layers (layers 4–6, 11–14, and post-21), suggesting that VP enhances cross-modal alignment at deeper stages of processing, particularly relevant to temporal reasoning.


\subsection{Details of the Analysis Subsets}
\Cref{tab:dataset_analysis} summarizes the evaluation subsets used in our analysis, where 100 examples were randomly sampled from each of the four datasets. Compared with the full temporal evaluation splits in \Cref{tab:dataset}, the sampled subsets show broadly similar distributions, especially in video length statistics across datasets and categories, indicating that the analysis subsets do not substantially deviate from the full benchmark distributions.

\section{Method Details}

\subsection{Sequential Visual Prompting}
%
\begin{figure}[!ht]
  \centering
  \includegraphics[width=\linewidth]{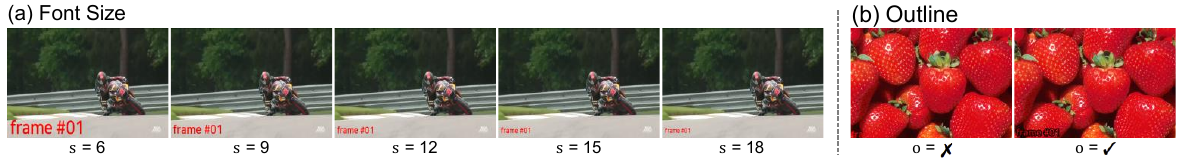}
  \vspace{-18 pt}
    \caption{\textbf{Effect of VP Design Parameters.}
    (a) Results with varying values of the font size control parameter $s$, and (b) results comparing the presence or absence of text outlines controlled by $o$.}
  \label{fig:fig_supp_vp}
\vspace{-10 pt}
\end{figure}
%

\paragraph{Implementation Details.}
We implement sequential visual prompting with a script that iterates over each video folder, sorts frames in chronological order, and overlays a text label of the form ``frame \#i'' on every frame at a fixed corner (TL, TR, BL, or BR). The font size is automatically scaled as $\text{fontsize} = \lfloor \min(\text{width}, \text{height}) / s \rfloor$, and the index is rendered in red text following prior works~\cite{VP,numberit}. The resulting VP-injected frames are shown in \cref{fig:fig_supp_vp}.

\subsection{Keyword-Frame Mapping (KFM)}
\begin{algorithm}[!ht]
\caption{Keyword--Frame Mapping (KFM)}
\label{alg:kfm}
\KwIn{user query $q$, sampled frames $\{\hat{\mathbf{I}}_i\}_{i=1}^{F}$, text tokens $\mathbf{Z}_{\text{text}}$, similarity threshold $\tau$}
\KwOut{augmented text tokens $\hat{\mathbf{Z}}_{\text{text}}$}

$\mathbf{W} \gets \mathrm{KeywordExtract}(q)$ \tcp*[r]{$\mathbf{W} = \{w_1, \dots, w_m\}$}
$\mathcal{K} \gets \emptyset$ \tcp*[r]{set of mapped frame indices}

\For{each keyword $w_j \in \mathbf{W}$}{
    \For{$i \gets 1$ \KwTo $F$}{
        $\mathbf{f}_i \gets \mathrm{CLIP}_{\text{img}}(\hat{\mathbf{I}}_i)$\;
        $\mathbf{g}_j \gets \mathrm{CLIP}_{\text{text}}(w_j)$\;
        $\mathrm{sim}_i \gets \cos(\mathbf{f}_i, \mathbf{g}_j)$\;
    }
    $k_j^\ast \gets \arg\max_i \mathrm{sim}_i$\;
    \If{$\mathrm{sim}_{k_j^\ast} \ge \tau$}{
        $\mathcal{K} \gets \mathcal{K} \cup \{k_j^\ast\}$ \tcp*[r]{map $w_j$ to frame $k_j^\ast$}
    }
}

$\hat{\mathbf{Z}}_{\text{text}} \gets \mathrm{InsertIndex}(\mathbf{Z}_{\text{text}}, \mathcal{K})$\;
\Return $\hat{\mathbf{Z}}_{\text{text}}$\;
\end{algorithm}


\noindent We extract keywords exclusively from the question text. In cases where a prompt contains both the question and the answer options in a single string, we first separate them and apply keyword extraction only to the question portion. This keeps the mapping focused on the actual information request of the prompt and avoids introducing option-specific cues into the extraction process.


\section{Additional Experiments}
\subsection{Additional Evaluation Results}

\subsubsection{Results on Non-Temporal Categories}


\begin{table}[!ht]
\centering
\caption{\textbf{Results on Non-Temporal Categories.}
Evaluation on the non-temporal categories of the VideoMME~\cite{videomme} dataset. The reported categories include: Action Reasoning (ARs), Action Recognition (ARc), Attribute Perception (AP), Counting Problem (CP), Information Synopsis (IS), OCR Problems (OP), Object Reasoning (ORs), Object Recognition (ORc), Spatial Perception (SP), Spatial Reasoning (SR), and Temporal Perception (TP). 
}
\resizebox{\linewidth}{!}{%

\begin{tabular}{
l!{\vrule}
>{\centering\arraybackslash}p{1.0cm}
>{\centering\arraybackslash}p{1.0cm}
>{\centering\arraybackslash}p{1.0cm}
>{\centering\arraybackslash}p{1.0cm}
>{\centering\arraybackslash}p{1.0cm}
>{\centering\arraybackslash}p{1.0cm}
>{\centering\arraybackslash}p{1.0cm}
>{\centering\arraybackslash}p{1.0cm}
>{\centering\arraybackslash}p{1.0cm}
>{\centering\arraybackslash}p{1.0cm}
>{\centering\arraybackslash}p{1.0cm}!{\vrule}
>{\centering\arraybackslash}p{1.0cm}
}
\toprule
Models & ARs & ARc & AP & CP & IS & OP & ORs & ORc & SP & SR & TP & mAcc \\
\hline

\multicolumn{13}{>{\columncolor{gray!20}}l}{\textit{64 Frames (1 FPS)}}\\

Baseline~\cite{llavavideo} & 
56.84 & 62.30 & 76.92 & 48.13 & 75.23 & 64.03 & 57.27 & 71.19 & 64.81 & 78.57 & 70.91 & 66.02 \\

\rowcolor{gray!10}
~+ Ours &
55.79 & 65.81 & 76.02 & 46.64 & 75.23 & 63.31 & 58.37 & 70.34 & 64.81 & 80.36 & 70.91 & 66.15~\textsubscript{\textcolor{red}{$\blacktriangle$}} \\
\hline

\multicolumn{13}{>{\columncolor{gray!20}}l}{\textit{20\% Frames}}\\

Baseline~\cite{llavavideo} & 
55.09 & 57.51 & 67.42 & 41.79 & 73.99 & 58.99 & 55.73 & 61.02 & 57.41 & 83.93 & 60.00 & 59.44 \\ 

\rowcolor{gray!10}
~+ Ours &
54.74 & 58.15 & 68.78 & 41.04 & 73.68 & 54.68 & 55.95 & 62.71 & 59.26 & 82.14 & 63.64 & 61.34~\textsubscript{\textcolor{red}{$\blacktriangle$}} \\

\bottomrule
\end{tabular}
}
\label{tab:tab_nontemporal_vidomme}
\end{table}


\noindent To examine how \ourmodel{} performs on a broader range of tasks beyond temporal reasoning, we report category-wise results on the general video understanding benchmark, VideoMME~\cite{videomme}. We observe that even for tasks not explicitly focused on temporal understanding, incorporating frame indices does not hinder performance and may offer auxiliary benefits for scene-level comprehension.

\subsubsection{Results Combined with the Keyframe Selection Method}

\begin{table}[!ht]
\centering
\caption{
\textbf{Integration with Keyframe Selection Methods.} Results showing the effect of combining our method with the state-of-the-art keyframe selection model AKS~\cite{aks} on long video datasets, VideoMME~\cite{videomme} and LongVideoBench~\cite{longvideobench}. This evaluates performance when applied to sampled frames from lengthy video inputs.
}
\vspace{-8pt}
\begin{tabular}{l!{\vrule}c!{\vrule}c!{\vrule}ccccc}
\toprule
\multirow{2}{*}{Models} 
& \multirow{2}{*}{Sampling} 
& \multicolumn{1}{c!{\vrule}}{VideoMME}
& \multicolumn{4}{c}{LongVideoBench} \\
& & TR & E3E & O3O & SSS & total \\
\hline

\multicolumn{7}{>{\columncolor{gray!20}}l}{\textit{64 Frames (1 FPS)}}\\
Baseline~\cite{llavavideo} & AKS~\cite{aks} & 51.41 & 67.02 & 57.58 & 41.24 & 54.86 \\
\rowcolor{gray!10}
~+ Ours & AKS~\cite{aks} &\textbf{53.67} & 72.34 & 57.58 & 39.18 & \textbf{56.03} \\
\hline\hline

\multicolumn{7}{>{\columncolor{gray!20}}l}{\textit{20\% Frames}}\\
Baseline~\cite{llavavideo} & AKS~\cite{aks} & 45.20 & 63.83 & 48.48 & 39.13 & 48.25 \\
\rowcolor{gray!10}
~+ Ours & AKS~\cite{aks} & \textbf{49.72} & 63.83 & 57.58 & 31.96 & \textbf{50.19} \\

\bottomrule
\end{tabular}%
\label{tab:table_AKS}
\end{table}

\noindent\Cref{tab:table_AKS} presents the results of applying keyframe-based sampling instead of uniform sampling in long-video scenarios. For this experiment, we used AKS~\cite{aks}, a state-of-the-art keyframe selection method. The results show a clear performance improvement over the baseline. This improvement likely stems from the increased inclusion of answer-relevant frames, particularly in the 64-frame setting, which raises the likelihood of retrieving frames aligned with extracted keywords and enables more accurate keyword-frame mapping. More broadly, this benefit can be extended to other keyframe selection methods~\cite{bolt,frame-voyager,quota}.

\subsubsection{Additional Performance Comparision with Token-Level Method}

\begin{table*}[!ht]
\centering
\caption{
\textbf{Comparision with Token-Level Pruning Method.} Results on the temporal understanding subsets of the TempCompass~\cite{tempcompass}, MVBench~\cite{mvbench}, VideoMME~\cite{videomme}, and LongVideoBench~\cite{longvideobench} datasets. 
}

\resizebox{\linewidth}{!}{%
\begin{tabular}{l!{\vrule}ccc!{\vrule}ccc!{\vrule}c!{\vrule}ccccc}
\toprule
\multirow{2}{*}{Models} 
& \multicolumn{3}{c!{\vrule}}{TempCompass}
& \multicolumn{3}{c!{\vrule}}{MVBench}
& \multicolumn{1}{c!{\vrule}}{VideoMME}
& \multicolumn{4}{c}{LongVideoBench}
\\
& EO & AC & total & AS & ST & total & TR & E3E & O3O & SSS & total \\
\hline

\multicolumn{12}{>{\columncolor{gray!20}}l}{\textit{30\% Frames (Tokens) }}\\

Baseline~\cite{llavavideo} & 71.24 & 69.32 & 70.27 & 67.00 & 90.00 & 78.50 & 44.63 & 63.83 & 45.45 & 36.08 & 48.64  \\
\rowcolor{gray!10}
~+ Ours & 73.63 & 71.95 & 72.78 & 67.50 & 93.00 & 80.25 & 50.28 & 64.89 & 46.97 & 38.14 & 50.19  \\
~+ Token Pruning ~\cite{framefusion} & 71.97 & 63.14 & 67.51 & 65.50 & 91.00 & 78.25 & 45.20 & 64.89 & 45.45 & 32.99 & 47.86  \\

\hline

\multicolumn{12}{>{\columncolor{gray!20}}l}{\textit{20\% Frames (Tokens) }}\\

Baseline~\cite{llavavideo} & 66.11 & 65.55 & 65.53 & 64.50 & 89.50 & 77.00 & 44.07 & 63.83 & 48.48 & 32.99 & 48.25 \\
\rowcolor{gray!10}
~+ Ours & 72.33 & 72.09 & 72.21 & 68.50 & 91.00 & 79.75 & 50.28 & 62.77 & 57.58 & 32.99 & 50.19 \\
~+ Token Pruning ~\cite{framefusion} & 68.14 & 58.10 & 63.07 & 58.50 & 91.50 & 75.00 & 45.20 & 63.83 & 42.42 & 32.99 & 46.69  \\

\bottomrule
\end{tabular}%
}
\label{tab:supple_efficiency}
\end{table*}

\noindent In Section 5.2 of the main paper, we compare our method against token-level pruning approaches under various frame-sampling ratios. Table~\ref{tab:supple_efficiency} shows the quantitative results for the 20\% and 30\% sampling settings. Overall, our approach consistently outperforms both the baseline and token-level methods across different sampling scenarios.

\subsubsection{Additional Qualitative Results}
\begin{figure}[!ht]
  \centering
  \includegraphics[width=\linewidth]{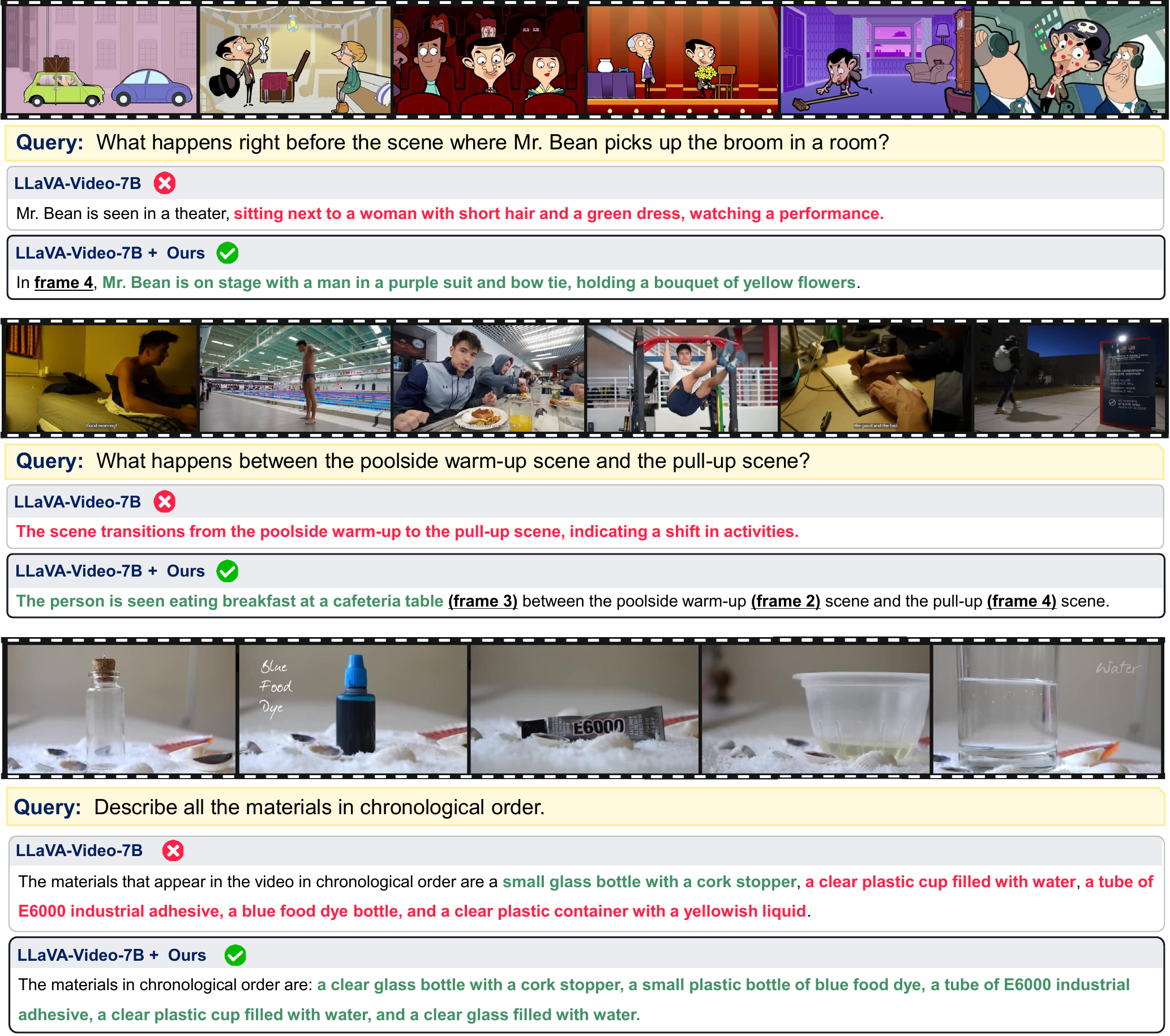}
  \vspace{-18 pt}
  \caption{
   \textbf{Additional qualitative temporal results.} Examples where our method provides temporally consistent answers, while the baseline VideoLLM fails to correctly capture the order of events.
   }
  \label{fig:fig_supp_qualitative}
\end{figure}

In \cref{fig:fig_supp_qualitative}, we illustrate how our methodology jointly
leverages frame-index visual prompting and keyword–frame mapping to improve
temporal reasoning. In the first example, the keyword \textit{``picks up the
broom''} is mapped to frame~5, allowing the model to correctly identify what
happens immediately before that moment. In the second example, the keywords
\textit{``poolside warm-up''} and \textit{``pull-up scene''} are aligned to
frames~2 and~4, respectively, enabling the model to reason about the events
that take place between these points. In these cases, queries are also augmented with the corresponding frame indices, such as
``What happens right before the scene where Mr. Bean picks up the broom
(frame~5) in a room?'' and ``What happens between the poolside warm-up
(frame~2) scene and the pull-up (frame~4) scene?'', which encourages the
model to explicitly reason over the numbered frames. In the third example, no
keyword is mapped to any frame, so the model relies solely on the injected
frame indices to recover the correct temporal order. These results highlight
that our methodology effectively combines both visual prompting and
keyword–frame correspondence to produce consistent temporal ordering across
diverse open-ended queries.

%
%

\subsection{Additional Ablation Results}

\subsubsection{Ablation on hyperparmeter $\tau$}

%
\begin{figure}[!ht]
  \centering
  \includegraphics[width=\linewidth]{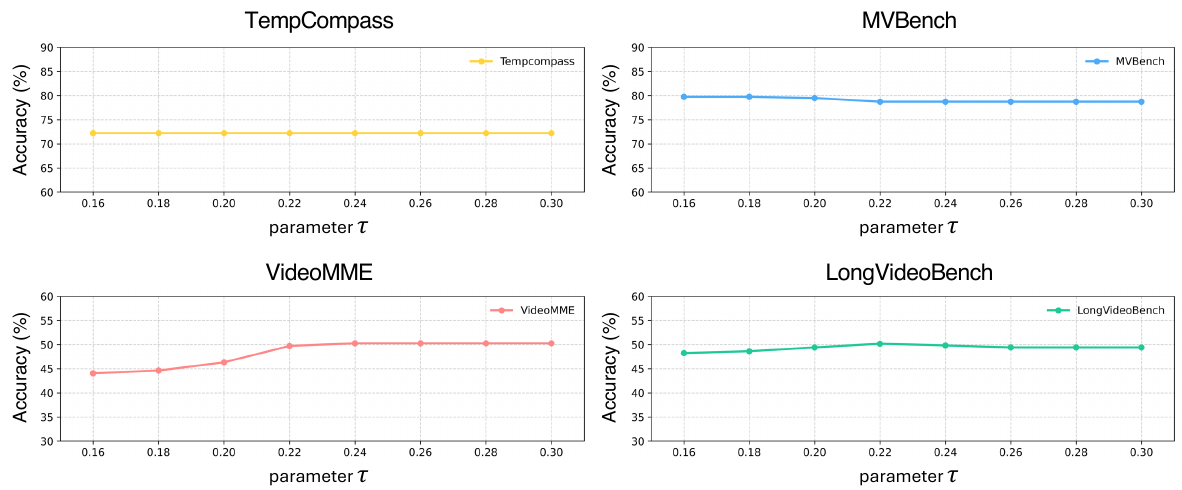}
  \vspace{-18 pt}
\caption{\textbf{Effect of Hyperparameter $\tau$.}
Performance variation across four datasets with different values of the mapping threshold $\tau$.}
  \label{fig:supple_ablation_t}
\vspace{-10 pt}
\end{figure}
%

\cref{fig:supple_ablation_t} illustrates the impact of varying the mapping threshold $\tau$ in the KFM module. When $\tau$ is set too low, irrelevant frames may be mapped to keywords, introducing noise and degrading performance. Conversely, a threshold that is too high reduces the number of successful mappings, thereby limiting the effectiveness of KFM. This trade-off often results in a single performance peak at an intermediate $\tau$ value, as observed in the graph. Notably, the TempCompass~\cite{tempcompass} dataset remains unaffected by changes in $\tau$, as its questions lack identifiable keywords, rendering the KFM module inactive and resulting in stable performance across all threshold settings.

\subsubsection{Ablation on hyperparmeter $s$}

%
\begin{figure}[!ht]
  \centering
  \includegraphics[width=\linewidth]{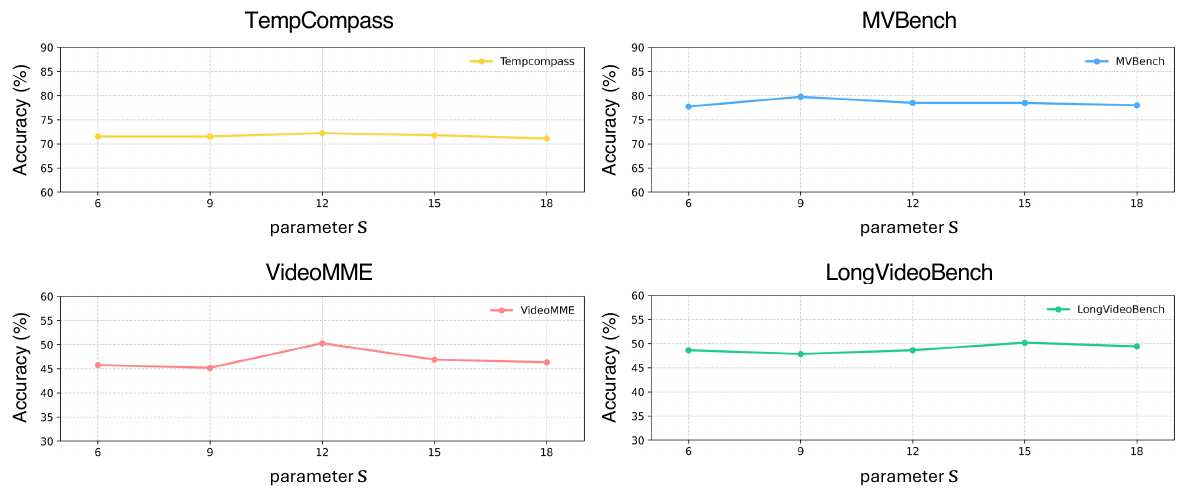}
  \vspace{-18 pt}
  \caption{\textbf{Effect of Hyperparameter $s$.}
Performance variation across four datasets as the font size control parameter $s$ changes.}
  \label{fig:fig_ablaion_s}
\vspace{-10 pt}
\end{figure}
%

\cref{fig:fig_ablaion_s}(a) illustrates how model performance varies with different values of the font size control parameter $s$. As shown in \cref{fig:fig_ablaion_s}(b), larger fonts improve the visibility of the frame numbers but may occlude key objects or text in the scene, potentially hindering recognition. Conversely, excessively small fonts are difficult to discern. These results suggest that the optimal font size is dataset-dependent, balancing visibility with minimal interference.

\subsubsection{Ablation on VP position}

%
\begin{figure}[!ht]
  \centering
  \includegraphics[width=\linewidth]{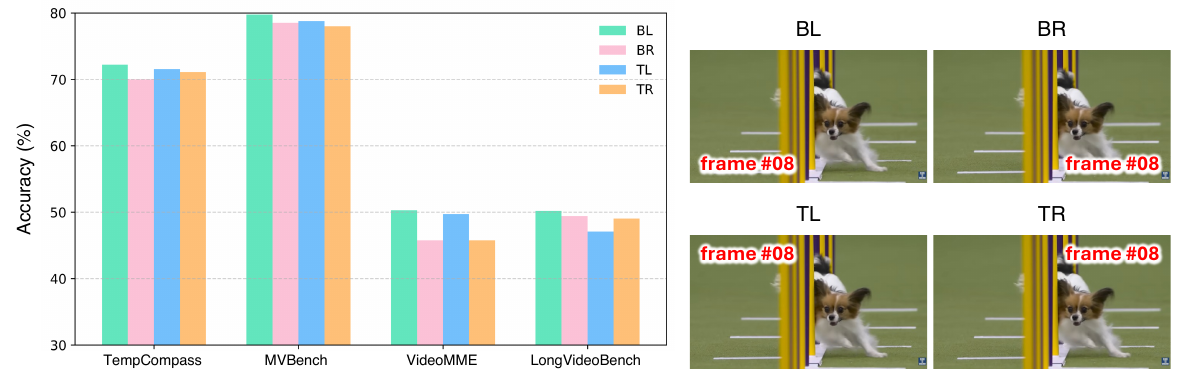}
  \vspace{-18 pt}
  \caption{\textbf{Effect of VP Position.}
Results on four datasets~\cite{tempcompass,mvbench,videomme,longvideobench} when the visual prompt is inserted at different frame positions: bottom-left (BL), bottom-right (BR), top-left (TL), and top-right (TR).}

  \label{fig:fig_ablation_position}
\vspace{-10 pt}
\end{figure}
%

\noindent
In Section 3.2 of the main paper, we observed that the VideoLLM exhibits a positional bias toward visual prompts (VP), which motivated our choice to place the VP at the bottom-left (BL) corner. \Cref{fig:fig_ablation_position} presents the results of inserting the VP at alternative positions. Consistent with our analysis, the BL placement yields the highest performance across all datasets.


\section{Implementation Details}

\subsection{Details of Evaluation Datasets}

\begin{table}[b!]
    \centering
    \caption{\textbf{Statistics of the temporal evaluation splits.} For each dataset, we report the number of questions, number of videos, and video length statistics (in seconds).}
    \vspace{2pt}
    \begin{tabular}{l c c c r r r r}
        \hlineB{2.5}

        Dataset & Category & Questions & Videos & Min (s) & Max (s) & Mean (s) & Std (s) \\
        \hline
        \multirow{3}{*}{TempCompass} 
            & EO    &    1384    &    100   &    3.04    &     29.36   &    13.98    &    6.67    \\
            & AC    &    1408    &   96    &   3.00     &   24.70     &    10.18    &    4.99    \\
            & Total &    2792    &    195   &   3.00     &    29.36    &    12.08    &    6.19    \\
        \hdashline 
        \multirow{3}{*}{MVBench} 
            & AS    &   200     &   179    &    7.10    &    43.20    &    20.57    &     7.93   \\
            & ST    &    200    &    200   &     20.00   &   20.00     &   20.00     &    0.00    \\
            & Total &   400     &    379   &    7.10    &    43.20    &    20.27    &   5.46     \\
        \hdashline
        \multirow{1}{*}{VideoMME} 
            & TR    &   177     &   171    &    56.89    &    3540.14    &   1459.26     &    1049.67    \\
        \hdashline
        \multirow{4}{*}{LongVideoBench} 
            & E3E   &   94     &   49    &   8.01     &  3485.12      &   587.69     &  719.64      \\
            & O3O   &   66     &   38    &    7.98    &   2041.01     &   404.31     &   521.95     \\
            & SSS   &   97     &   52    &   8.57     &   1995.88     &   469.75     &   501.24     \\
            & Total &   257     &   139    &   7.98     &   3485.12     &   493.44     &   597.11     \\
        \hlineB{2.5}
    \end{tabular}
\label{tab:dataset}
\end{table}

Our method works reliably across a broad range of temporal reasoning tasks, but is particularly effective in scenarios with frequent scene transitions or questions about what happens before or after specific moments. For this reason, we focus on benchmark categories that require explicit reasoning over event and scene changes, such as the temporal subsets of TempCompass and MVBench. To verify that these gains also hold in more complex and long-form settings, we additionally evaluate on the temporal reasoning splits of VideoMME and LongVideoBench.

\paragraph{TempCompass.}
TempCompass~\cite{tempcompass} is a diagnostic benchmark designed to probe fine-grained temporal perception in Video LLMs across diverse temporal aspects and task formats. It contains diverse videos paired with four query types: caption matching, captioning, multiple-choice QA, and yes/no QA. In our experiments, we use only the temporal reasoning subset, namely the Event Order (EO) and Attribute Change (AC) categories. EO evaluates whether the model can infer the chronological order of events, while AC tests whether it can track how object attributes or global scene states evolve over time.

\paragraph{MVBench.} 
MVBench~\cite{mvbench} is a comprehensive multi-modal video understanding benchmark that evaluates temporal skills beyond what can be captured from single static frames. It covers 20 video tasks derived from image-based benchmarks and converted into temporal, video-level multiple-choice QA. In our experiments, we use only the temporal reasoning subset, namely the Action Sequence (AS) and Scene Transition (ST) categories. AS evaluates whether the model can retrieve the events occurring before or after a specific action, while ST tests whether it can determine how the scene transitions over the course of the video.

\paragraph{VideoMME.}
VideoMME~\cite{videomme} is a large-scale video benchmark designed to evaluate multimodal LLMs on fine-grained, video-grounded question answering across diverse domains. It includes a dedicated temporal reasoning (TR) subset, where questions require understanding event progression, temporal relations, and dependencies over extended video clips rather than single frames. In our evaluation, we use only this TR subset to specifically assess how well models can reason about when events occur and how they unfold over time.

\paragraph{LongVideoBench.}
LongVideoBench~\cite{longvideobench} is a long-form video question answering benchmark with video–language interleaved inputs of up to an hour. In our evaluation, we use the temporal reasoning subsets E3E, O3O, and SSS, which focus on before/after relations between events, the order of entities or concepts, and the correct sequence of scenes across the video.

\subsection{Hyperparameter Settings}
\label{sec:Hyperparameter_settings}

\begin{table*}[t]
\centering
\caption{\textbf{Hyperparameter Settings.}  
Hyperparameters selected for four different VideoLLM baselines~\cite{gpt4_1,qwen2vl,llavavideo,llavaonevision} across four datasets~\cite{tempcompass,mvbench,videomme,longvideobench}, under two frame sampling conditions (64 frames and 20\% frames), as reported in Table~2 of the main paper. Notation is provided Section~\ref{sec:Hyperparameter_settings}.
}
\resizebox{\linewidth}{!}{%

\begin{tabular}{
l!{\vrule}
>{\centering\arraybackslash}p{1.0cm}
>{\centering\arraybackslash}p{1.0cm}
>{\centering\arraybackslash}p{1.0cm}!{\vrule}
>{\centering\arraybackslash}p{1.0cm}
>{\centering\arraybackslash}p{1.0cm}
>{\centering\arraybackslash}p{1.0cm}!{\vrule}
>{\centering\arraybackslash}p{1.0cm}
>{\centering\arraybackslash}p{1.0cm}
>{\centering\arraybackslash}p{1.0cm}!{\vrule}
>{\centering\arraybackslash}p{1.0cm}
>{\centering\arraybackslash}p{1.0cm}
>{\centering\arraybackslash}p{1.0cm}
}
\toprule
\multirow{2}{*}{Models} 
& \multicolumn{3}{c!{\vrule}}{TempCompass}
& \multicolumn{3}{c!{\vrule}}{MVBench}
& \multicolumn{3}{c!{\vrule}}{VideoMME}
& \multicolumn{3}{c}{LongVideoBench}
\\
& $\tau$ & $s$ & $o$ & $\tau$ & $s$ & $o$ & $\tau$ & $s$ & $o$ & $\tau$ & $s$ & $o$ \\
\hline

\multicolumn{13}{>{\columncolor{gray!20}}l}{\textit{64 Frames (1 FPS)}}\\

Ours~\textsubscript{\textit{GPT4.1}} & 
1.00 & 12 & \myxmark & 
0.18 & 9 & \myxmark &
- & - & - &
- & - & - \\

Ours~\textsubscript{\textit{QwenVL}} & 
1.00 & 12 & \myxmark & 
0.20 & 9 & \myxmark &
0.15 & 12 & \myxmark &
0.15 & 15 & \myxmark  \\

Ours~\textsubscript{\textit{LLaVA-Video}} &
1.00 & 12 & \myxmark & 
0.18 & 9 & \myxmark &
0.23 & 12 & \myxmark &
0.22 & 12 & \mychecking \\

Ours~\textsubscript{\textit{LLaVA-OneVision}} & 
1.00 & 12 & \myxmark & 
0.18 & 9 & \mychecking &
0.28 & 15 & \mychecking &
0.18 & 12 & \mychecking \\
\hline

\multicolumn{13}{>{\columncolor{gray!20}}l}{\textit{20\% Frames}}\\

Ours~\textsubscript{\textit{GPT4.1}} & 
1.00 & 12 & \myxmark & 
0.18 & 9 & \myxmark &
- & - & - &
- & - & - \\

Ours~\textsubscript{\textit{QwenVL}} & 
1.00 & 12 & \myxmark & 
0.20 & 9 & \myxmark &
0.32 & 12 & \myxmark &
0.22 & 15 & \myxmark \\

Ours~\textsubscript{\textit{LLaVA-Video}} &
1.00 & 12 & \myxmark & 
0.18 & 9 & \myxmark &
0.24 & 12 & \myxmark &
0.22 & 16 & \myxmark \\

Ours~\textsubscript{\textit{LLaVA-OneVision}} &
1.00 & 12 & \myxmark & 
0.20 & 12 & \mychecking &
0.27 & 9 & \mychecking &
0.24 & 12 & \mychecking \\
\bottomrule
\end{tabular}
}
\label{tab:tab_hyperparameters}
\end{table*}

\noindent We apply optimized values for three key hyperparameters based on the dataset, model size, and number of frames used in each experimental setting, as summarized in \Cref{tab:tab_hyperparameters}.

\paragraph{Font Size Controller $s$.}
The parameter $s$, introduced in Section~4.2 of the main paper, controls the relative font size of the visual prompt (VP) text with respect to the resolution of each video frame. This ensures that the VP is clearly visible regardless of frame size or aspect ratio.

\paragraph{Mapping Threshold $\tau$.}
The threshold $\tau$, as defined in Section~4.3 of the main paper, determines whether a keyword-to-frame mapping is applied during the KFM process. Mapping is only performed if the frame with the highest similarity to a given keyword exceeds this threshold, thereby preventing incorrect associations with semantically irrelevant frames.
This condition assumes that meaningful keywords can be extracted from the user query. In the case of the TempCompass~\cite{tempcompass} dataset, however, keyword extraction did not yield any informative terms from the question (e.g., \textit{“Which scene is the most suspenseful?”}). As a result, the mapping mechanism was effectively bypassed, and we set $\tau = 1.0$ uniformly for all instances in this dataset. For all other benchmarks, keywords are extracted only from the question portion of the user query, excluding any answer options or candidates.

\paragraph{Outline Usage $o$.}
The parameter $o$ is a boolean flag that determines whether an outline is applied to the VP text to enhance visibility. This feature was introduced to improve legibility in cases where red-colored VP text overlaps with red or visually complex regions within a frame. In datasets such as VideoMME~\cite{videomme} and LongVideoBench~\cite{longvideobench}, which contain relatively cluttered scenes, the default setting without outlines often led to poor visibility. In addition, recognition performance varied depending on the specific VideoLLM model, particularly in models like LLaVA-OneVision~\cite{llavaonevision}, so the outline option was selectively used to improve visual clarity.

\subsection{Prompt Templates}

\subsubsection{System Prompt Templates}
We adopt the default system instruction (\texttt{You are a helpful assistant.}), and extend it with an additional guideline that encourages temporal grounding. Specifically, we append the following sentence to the instruction: \texttt{Focus on the temporal relationships by referring to the number written in the \{position\} corner \\ of each frame.} 
This prompt steers the model to explicitly use the injected visual prompts when interpreting temporal relationships across frames.

\subsubsection{User Prompt Templates}
User prompts follow the standard formatting used in LMMs-Eval~\cite{lmms_eval2024}, ensuring that each model receives questions in a consistent and comparable manner. Detailed examples of the user prompts can be found in \Cref{tab:supp_prompts}.

\begin{table*}[ht!]
\centering
\small
\begin{tabular}{p{0.18\linewidth} p{0.76\linewidth}}
\toprule
\textbf{Role / Dataset} & \textbf{Prompt} \\
\midrule
\rowcolor{gray!10}\multicolumn{2}{l}{\textbf{System (shared across all datasets)}}\\
\texttt{system} &
\begin{minipage}[t]{\linewidth}\ttfamily\small
You are a helpful assistant.\\
Focus on the temporal relationships by referring to the number 
written in the bottom-left corner of each frame.
\end{minipage}
\\
\midrule
\rowcolor{gray!10}\multicolumn{2}{l}{\textbf{TempCompass}}\\
\texttt{user} &
\begin{minipage}[t]{\linewidth}\ttfamily\small
\$\{question\}\\
\$\{options\}\\[0.4em]
Task-specific post-prompt:\\
- Multi-choice: 'Please directly give the best option.'\\
- Yes/No: 'Please answer yes or no.'\\
- Caption matching: 'Please directly give the best option.'\\
- Captioning: 'Please directly give the best option.'
\end{minipage}
\\[1.0em]
\textbf{Example} &
\begin{minipage}[t]{\linewidth}\ttfamily\small
Which event happens first to the skillet?\\
A. Burning in fire\\
B. None of both\\
C. Smoking\\[0.4em]
Please directly give the best option.
\end{minipage}
\\
\midrule
\rowcolor{gray!10}\multicolumn{2}{l}{\textbf{MVBench}}\\
\texttt{user} &
\begin{minipage}[t]{\linewidth}\ttfamily\small
\$\{question\}\\
\$\{options\}\\[0.4em]
Only give the best option.
\end{minipage}
\\[1.0em]
\textbf{Example} &
\begin{minipage}[t]{\linewidth}\ttfamily\small
What happened after the person took the food?\\
A. Ate the medicine.\\
B. Tidied up the blanket.\\
C. Put down the cup/glass/bottle.\\
D. Took the box.\\[0.4em]
Only give the best option.
\end{minipage}
\\
\midrule
\rowcolor{gray!10}\multicolumn{2}{l}{\textbf{VideoMME}}\\
\texttt{user} &
\begin{minipage}[t]{\linewidth}\ttfamily\small
\$\{question\}\\
\$\{options\}\\[0.4em]
The best answer is:
\end{minipage}
\\[1.0em]
\textbf{Example} &
\begin{minipage}[t]{\linewidth}\ttfamily\small
What kind of communication is listed before Semaphore?\\
A. Telephone.\\
B. Homing pigeon.\\
C. Telegraph.\\
D. Pony express.\\[0.4em]
The best answer is:
\end{minipage}
\\
\midrule
\rowcolor{gray!10}\multicolumn{2}{l}{\textbf{LongVideoBench}}\\
\texttt{user} &
\begin{minipage}[t]{\linewidth}\ttfamily\small
\$\{question\}\\
\$\{options\}\\[0.4em]
Answer with the option's letter from the given choices directly.
\end{minipage}
\\[1.0em]
\textbf{Example} &
\begin{minipage}[t]{\linewidth}\ttfamily\small
What is the color of the first piece of clothing shown in the video?\\
A. white\\
B. purple\\
C. red\\
D. olive\\
E. black\\[0.4em]
Answer with the option's letter from the given choices directly.
\end{minipage}
\\
\bottomrule
\end{tabular}
\caption{System prompt and dataset-specific user prompt templates with examples.}
\label{tab:supp_prompts}
\end{table*}

\subsubsection{Keyword Extractor Prompt Templates}
We extract text keywords from user queries using a unified LLM-based procedure across all benchmarks. 
The extractor uses Qwen2.5-7B-Instruct with a fixed system prompt and dataset-specific instruction templates. 
In particular, we employ a single system prompt that enforces keyword-only outputs and a shared user prompt that specifies generic extraction rules. 
On top of this shared template, we attach short dataset-specific user instructions and examples that mirror the question style of each benchmark.
This allows the extractor to adapt to different temporal reasoning formats while preserving a unified keyword output, as summarized in \Cref{tab:supp_keyword_prompts}.

\begin{table*}[ht!]
\centering
\small
\begin{tabular}{p{0.18\linewidth} p{0.76\linewidth}}
\toprule
\textbf{Role / Component} & \textbf{Prompt} \\ 
\midrule

\rowcolor{gray!10}\multicolumn{2}{l}{\textbf{System Prompt (Shared)}}\\
\texttt{system} &
\begin{minipage}[t]{\linewidth}\ttfamily\small
You are a helpful assistant that only extracts keywords and outputs them as a Python list.
\end{minipage}
\\
\midrule

\rowcolor{gray!10}\multicolumn{2}{l}{\textbf{User Prompt (Shared Core Instructions)}}\\
\texttt{user} &
\begin{minipage}[t]{\linewidth}\ttfamily\small
Follow these rules carefully:\\
1. Identify Key Phrases: Your goal is to extract key phrases from the question that refer to specific scenes, events, actions, or distinct items. \\
2. Exact Extraction: The extracted phrases must appear exactly as they do in the question. Do not modify or rephrase them.\\
3. Empty List Condition: If no relevant key phrases (as defined in Rule 1) are found in the question, you must return an empty list [].\\[0.4em]
Now:\\
Question: \{\$question\}\\
Your Answer:
\end{minipage}
\\
\midrule

\rowcolor{gray!10}\multicolumn{2}{l}{\textbf{Dataset-Specific User Prompt Examples}}\\

\textbf{TempCompass} &
\begin{minipage}[t]{\linewidth}\ttfamily\small
Example 1:\\
Question: Which sentence better captures the essence of the video?\\
Your Answer: []\\[0.4em]
Example 2:\\
Question: Which description is a more suitable match for the video?\\
Your Answer: []
\end{minipage}
\\
\midrule

\textbf{MVBench} &
\begin{minipage}[t]{\linewidth}\ttfamily\small
Example 1:\\
Question: What happened after the person took the food?\\
Your Answer: ["the person took the food"]\\[0.4em]
Example 2:\\
Question: What happened after the person closed the door?\\
Your Answer: ["the person closed the door"]
\end{minipage}
\\
\midrule

\textbf{VideoMME} &
\begin{minipage}[t]{\linewidth}\ttfamily\small
Example 1:\\
Question: When is the zodiacal light visible from the video? \\
Your Answer: ["the zodiacal light"]\\[0.4em]
Example 2:\\
Question: Which GPT is introduced after Convert Anything?\\
Your Answer: ["Convert Anything"]
\end{minipage}
\\
\midrule

\textbf{LongVideoBench} &
\begin{minipage}[t]{\linewidth}\ttfamily\small
Example 1:\\
Question: In front of a blue background, a gentleman wearing a shirt with pink floral patterns is speaking. What did the gentleman do after becoming friends with the unicorn?\\
Your Answer: ["gentleman wearing a shirt with pink floral patterns is speaking", "becoming friends with the unicorn"]\\[0.4em]
Example 2:\\
Question: In the movie scene, there is a man in gray-black clothes standing between a red door and wall on the left, and a silver-white window and yellow wall on the right. After this man appears, which person or object appears first?\\
Your Answer: ["man in gray-black clothes standing", "a red door and wall on the left, and a silver-white window and yellow wall on the right"]
\end{minipage}
\\

\bottomrule
\end{tabular}
\caption{System prompt, shared keyword-extraction instructions, and dataset-specific examples used for keyword extraction.}
\label{tab:supp_keyword_prompts}
\end{table*}

\clearpage
\section{Discussion}

\subsection{Clarification on Novelty}
Prior works like Number it~\cite{numberit} utilize frame numbering for temporal grounding.  That is, numbers recognized via OCR serve as frame indices to guide output formats and mitigate hallucinations. In contrast, we use VP for temporal reasoning, where the model needs to understand multiple events, infer their temporal relations, and derive high-level reasoning. Thus, simply treating VP as frame indices is insufficient for such complex reasoning tasks. Within our framework, VP is redefined as \textit{dictionary keys} for our KFM, precisely aligning textual concepts with relevant frames. We note that our work is the first to identify and methodologically utilize VP as strong temporal cues for various complex temporal reasoning tasks.

\begin{table}[t]
\centering

\setlength{\tabcolsep}{3pt}
\setlength{\columnsep}{6pt}
\setlength{\aboverulesep}{1pt}
\setlength{\belowrulesep}{1pt}
\setlength{\cmidrulesep}{1pt}
\renewcommand{\arraystretch}{0.8}

\begin{minipage}[t]{0.48\linewidth}
\centering
\resizebox{\linewidth}{!}{
\begin{tabular}{l c c c c}
\toprule
    & \multicolumn{2}{c}{\small VideoMME} & \multicolumn{2}{c}{\small LongVideoBench} \\
    & 1fps & 20\% & 1fps & 20\% \\
    \midrule
    Baseline~\cite{llavavideo} & 47.46 & 44.07 & 56.42 & 48.25 \\
    Interleaved Text Token & 0.00 & 44.07 & 22.96 & 47.86 \\
    Structured Timeline & 47.46 & 45.76 & 55.64 & 46.69 \\
    Random Number VP & 46.33 & 44.07 & 54.09 & 47.85 \\
    \rowcolor{gray!10}
    Ours & 48.02 & 50.28 & 56.42 & 50.19 \\
\bottomrule
\end{tabular}
}
\caption{Comparison with alternative prompting strategies.}
\label{tab:prompt_compare}
\end{minipage}
\hfill
\begin{minipage}[t]{0.48\linewidth}
\centering
\resizebox{\linewidth}{!}{
\begin{tabular}{l c c c c}
\toprule
    & \small 1fps & \multicolumn{3}{c}{\small 20\%} \\
    \cmidrule(r){2-2} \cmidrule(l){3-5}
    & \small Baseline~\cite{llavavideo} & \small Baseline~\cite{llavavideo} & \small Ours (w/o KFM) & \small Ours \\
    \midrule
    FLOPs ($\times 10^{13}$) & 5.47 & 2.05 & 2.07 & 2.08 \\
    Latency (sec) & 12.86 & 1.53 & 1.63 & 2.31 \\
\bottomrule
\end{tabular}
}
\caption{Computation cost under different settings.}
\label{tab:cost_compare}
\end{minipage}

\end{table}
\subsection{Comparison with Alternative Prompting Methods}
We compare ViKey with several alternative prompting strategies, including Interleaved Text Token, Structured Timeline, and Random Number VP. As shown in \Cref{tab:prompt_compare}, sequential visual prompts consistently achieve the best performance under sparse-frame settings across both VideoMME and LongVideoBench. While some alternatives provide minor improvements over the baseline in specific cases, none outperform ViKey overall. In particular, text interleaving often leads to unstable behavior and incoherent outputs as the number of frames increases. These results support the use of sequential visual prompts as an effective and practical plug-and-play design for pretrained VideoLLMs.

\subsection{Computational Cost}
We further report the end-to-end computational cost in \cref{tab:cost_compare}. Compared with the dense 1\,fps baseline, our sparse setting substantially reduces FLOPs and latency, while adding KFM introduces only a small additional cost over the VP-only variant. Since the method is training-free, it also avoids any extra training overhead.

\subsection{Occlusion Concerns}
\begin{wrapfigure}{r}{0.4\linewidth} 
    \centering
    \vspace{-25pt}
    \includegraphics[width=\linewidth]{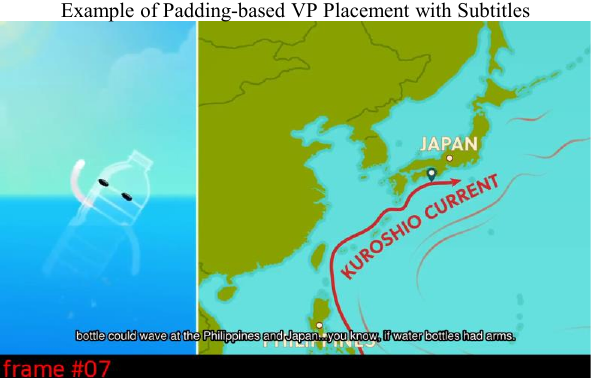}
   \caption{
    \textbf{Flexible zero-padding to reduce overlap with on-screen text.} This helps reduce interference with subtitles and task-relevant content.}
    \vspace{-12pt} 
    \label{fig:occlusion_pad}
\end{wrapfigure}
Because \ourmodel{} inserts frame indices directly into the visual input, an important question is whether the prompt may interfere with pre-existing textual elements or partially occlude task-relevant content. To examine this issue, we evaluated 163 VideoMME (TR) clips containing explicitly encoded subtitles. In this setting, the baseline achieves 39.88\% accuracy, while our default VP configuration reaches 42.94\%(+3.06), indicating that the added visual prompts remain effective even in the presence of real-world text.
To further reduce possible overlap, we additionally consider a flexible zero-padding configuration, illustrated on the right. This variant improves accuracy to 44.17\%(+4.29), suggesting that simple layout adjustments can further mitigate interference between the visual prompt and existing on-screen content.

\subsection{Limitations}
\begin{figure}[!t]
  \centering
  \includegraphics[width=\linewidth]{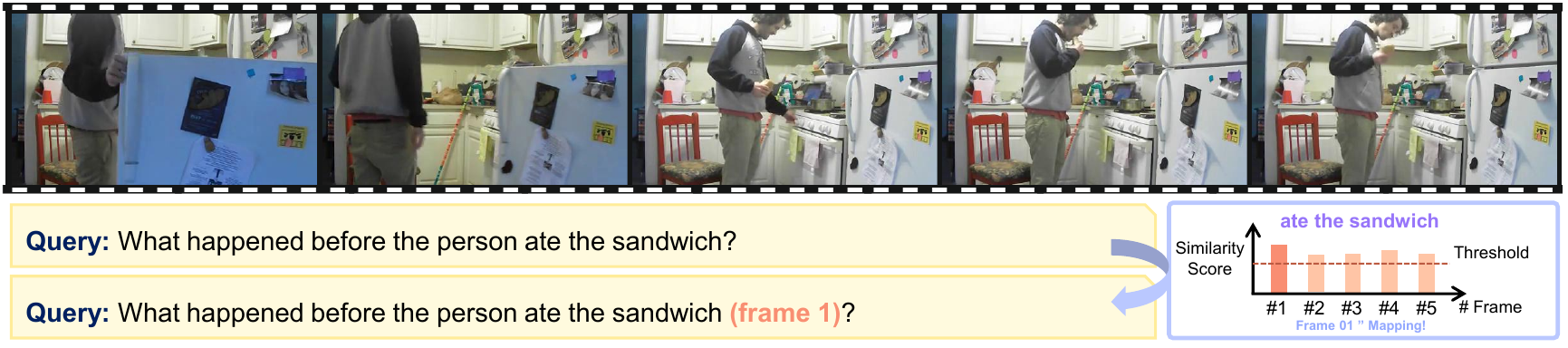}
  \vspace{-18 pt}
  \caption{
   \textbf{Example of ambiguous keyword–frame mapping.} When a video contains many visually similar consecutive frames, CLIP-based similarity may not be sufficient for accurate mapping.
   }
  \label{fig:fig_supp_failure}
\vspace{-10 pt}
\end{figure}

Although Keyword-to-Frame Mapping (KFM) generally aligns textual cues with the correct temporal moment, we observe that CLIP-based similarity can be ambiguous when consecutive frames are visually near-identical, as illustrated in \cref{fig:fig_supp_failure}. In such cases, multiple adjacent frames obtain nearly the same similarity score, causing the model to map the keyword (e.g., ate the sandwich) to an incorrect frame. Because the query is then augmented with this erroneous frame index, the model may reason around the wrong temporal anchor, leading to subtle but consistent mistakes in temporal reasoning. In addition, because \ourmodel{} inserts the frame index at a fixed corner of each frame, the visual prompt can occasionally occlude task-relevant content. Although the injected index occupies only a small region, certain videos place important objects or discriminative cues near the frame corners. When this occurs, the prompt may partially cover these regions and alter the appearance of the underlying content, which can in turn lead the model to misinterpret the scene or miss fine-grained details.

\subsection{Future Works}
Building on these observations, several extensions could further improve the stability and robustness of \ourmodel{}.
First, to mitigate the ambiguity observed when consecutive frames are visually near-identical, a promising direction is to incorporate sequence-aware or multi-frame similarity, where keyword matching considers short temporal windows rather than individual frames.
Leveraging temporal gradients, motion cues, or lightweight event-boundary detectors may also help distinguish frames that are visually similar but temporally distinct, reducing the likelihood of attaching a keyword to a neighboring but incorrect frame.
Second, to address the occasional occlusion caused by placing the prompt at a fixed corner, future work may explore adaptive prompt placement strategies. Simple heuristics based on visual saliency, object occupancy, or corner-level feature density could identify unobtrusive regions for inserting the frame index.
Together, these extensions would preserve the simplicity of \ourmodel{} while improving its reliability across diverse video layouts and scene structures.

%
%

\end{document}